\documentclass[10pt,journal]{IEEEtran}
\usepackage{setspace}

\usepackage[margin=0.60in]{geometry}

\usepackage[utf8]{inputenc} % allow utf-8 input
\usepackage[T1]{fontenc}    % use 8-bit T1 fonts
\usepackage{hyperref}       % hyperlinks
\usepackage{url}            % simple URL typesetting
\usepackage{booktabs}       % professional-quality tables
\usepackage{amsfonts}       % blackboard math symbols
\usepackage{nicefrac}       % compact symbols for 1/2, etc.
\usepackage{microtype}      % microtypography
\usepackage{lipsum}
\usepackage{graphicx}
\usepackage{amsmath} 
\usepackage[ruled,vlined]{algorithm2e}      % blackboard math symbols
\usepackage{amsthm}
\usepackage{subcaption}
\usepackage[noadjust]{cite}
\usepackage[absolute,overlay]{textpos}
\pdfoutput=1

\ifCLASSINFOpdf

\else

\fi

\LinesNumbered

\allowdisplaybreaks
\begin{document}

\begin{textblock*}{19cm}(0.3cm,0.3cm) % {block width} (coords) 
	\small
\noindent Accepted to be published in IEEE TRANSACTIONS ON COMMUNICATIONS in 2021. IEEE Copyright Notice: \copyright 2021 IEEE. Personal use of this material is permitted. Permission from IEEE must be obtained for all other uses, in any current or future media, including reprinting/republishing this material for advertising or promotional purposes, creating new collective works, for resale or redistribution to servers or lists, or reuse of any copyrighted component of this work in other works. 
\end{textblock*}

\title{Fast Federated Learning by Balancing Communication Trade-Offs}

\author{Milad Khademi Nori, Sangseok Yun, and Il-Min Kim, {\it Senior Member, IEEE} 
	%	\thanks{Manuscript was submitted on September 25, 2020.}
	\thanks{Milad Khademi Nori and Il-Min Kim are with the Department of Electrical and Computer Engineering, Queen's University, Kingston, ON K7L 3N6, CA (e-mail: 19mkn1@queensu.ca; ilmin.kim@queensu.ca). Sangseok Yun is with the Department of Information and Communications Engineering, Pukyong National University, Busan 48513, South Korea (e-mail: ssyun@pknu.ac.kr).}
}

\maketitle

\begin{abstract}
Federated Learning (FL) has recently received a lot of attention for large-scale privacy-preserving machine learning. However, high communication overheads due to frequent gradient transmissions decelerate FL.
To mitigate the communication overheads, two main techniques have been studied: (i) local update of weights characterizing the trade-off between communication and computation and (ii) gradient compression characterizing the trade-off between communication and precision. To the best of our knowledge, studying and balancing those two trade-offs \textit{jointly and dynamically} while considering their impacts on convergence has remained unresolved even though it promises significantly faster FL.
In this paper, we first formulate our problem to minimize learning error with respect to two variables: \textit{local update coefficients} and \textit{sparsity budgets} of gradient compression who characterize trade-offs between communication and computation/precision, respectively. We then derive an upper bound of the learning error in a given wall-clock time considering the interdependency between the two variables. Based on this theoretical analysis, we propose an enhanced FL scheme, namely Fast FL (FFL), that jointly and dynamically adjusts the two variables to minimize the learning error.
We demonstrate that FFL consistently achieves higher accuracies faster than similar schemes existing in the literature.
\end{abstract}


\IEEEpeerreviewmaketitle

\section{Introduction} 

\IEEEPARstart{F}{ederated} 
Learning (FL) is a training paradigm where a large number of workers collectively train a model using Stochastic Gradient Descent (SGD) \cite{1stPar1st}. Each worker holds a local (often private) training dataset, with which it contributes to training a global model. Specifically, in each FL round, each worker computes the error of the global model by forward-propagating the samples of its own dataset through the model, followed by comparing the model's outputs with samples' labels and applying an appropriate loss function. By backpropagating this error, each worker calculates the gradients. All workers then send their gradients to the server, who updates the global model and sends the updated model back to all the workers. At this point, one FL round is completed. After a number of (FL) rounds, the global model converges \cite{1stPar3rd}.

FL has recently become popular for the following reasons: (i) FL enables privacy-preserving training as workers only share their gradients with the server. That is, FL makes it possible to train a model from the entire datasets of workers without any individuals having to reveal their local datasets to other workers or the server. This characteristic of FL technology renders it apposite to many applications, the most popular of which is Internet of Things (IoT) where the data from organizations, hospitals, homes (Smart Home), and vehicles (Internet of Vehicles) presumably needs to be collected for processing, inference, and decision-making \cite{2ndPar1st, 2ndPar2nd}. FL removes the need to collect the raw data in using machine learning for IoT, thereby it ameliorates the privacy of IoT. (ii) FL is an exemplar of Edge Intelligence, which is the offspring from the union of (mobile) Edge Computing and Artificial Intelligence (AI). Edge Intelligence advocates for pushing the intelligence down from clouds to the proximity of end-users by implementation of AI applications on edge-devices\footnote{While some papers in literature of FL use the term ``edge-device'' as to refer to the nodes at which the gradients are calculated, in this paper, we prefer to use the term ``worker.''}, thereby promising better privacy and reliability with lower latency and cost. FL, correspondingly, can be used to facilitate training of large-scale models by parallelizing the computation of gradients on edge-devices. Moreover, FL allows inference to be performed on edge-devices, and thus, fulfills the goals of Edge Intelligence \cite{wang2020convergence}. (iii) FL addresses the intelligence democratization issue: in the AI market, big companies possessing a large amount of data plus computational and storage facilities can get the monopoly of AI businesses while small firms have no chances in this competition. FL enables small businesses to play a role in the market as they do not require a massive amount of data at a datacenter, nor computational and storage facilities.

Unfortunately, the empirical speedup (for training) offered by FL often fails to meet the optimal scaling (in the number of workers) that is ideally desired. It is now widely acknowledged that this speedup saturation is mainly due to the communication overheads, which are largely attributed to frequent gradient transmissions from workers to the server. As the number of parameters in the state-of-the-art models scales to a huge number (e.g., hundreds of millions), the size of the gradients scales proportionally. The communication bottleneck becomes even more pronounced particularly when the workers performing FL are wireless devices (e.g., smartphones and sensors) which communicate through wireless channels and suffer from low-bandwidth, intermittent connections, and expensive mobile data plans \cite{theRecentSurvey, local, adacom, adaptau}. To mitigate the communication overhead problem and hence speedup learning, in the literature, two main techniques have been studied. The first is local update, characterizing the trade-off between communication and computation, and the second is gradient compression, characterizing the trade-off between communication and precision. Each of these techniques will be discussed in the following.

In the local update technique, within each round, many local updates are performed---instead of only one update. Local update presents a communication-computation trade-off as \textit{local update coefficient} determines the ratio of computation to communication. The authors of \cite{local, adacom, adaptau} showed that local updates can significantly alleviate frequent gradient transmissions. Formulating the error-convergence bound in terms of local updates, Adaptive Communication (ADACOMM) \cite{adacom} dynamically optimized the local update coefficients to accelerate convergence while retaining accuracy. ADACOMM also showed that the communication-computation trade-off introduced by local update coefficients should be dynamically balanced over the course of learning. In \cite{adapjsac}, the same issue was addressed, but assuming that the loss function was convex. Performing binary compression and benefiting from local updates, Sparse Binary Compression (SBC) \cite{sbc} could significantly decrease communication overheads. 

Gradient compression presents the communication-precision trade-off, i.e., how much compression is productive in different rounds? Gradient compression methods fall into two categories: gradient quantization and sparsification. Gradient quantization approaches the problem of communication overhead by sending a compressed (quantized) version of the gradient. TernGrad \cite{ternGrad} used 2 bits for each element of the gradient and authors mathematically proved the convergence of TernGrad under the assumption of a bound on gradients. Quantized SGD (QSGD) \cite{qsgd} investigated the effects of 2 bits, 4 bits, and 8 bits quantizations on different layers of neural networks and showed that the networks converge. Authors of \cite{one} even went further and quantized the gradients to one bit. In \cite{quan1} and \cite{quan2}, similar quantization techniques were exploited.

Sparsification technique aims to sparsify gradients in order to send only the significant elements of the gradients instead of all. Sparsification techniques fall into two categories depending on in what domain the sparsity is sought for. Some researchers assume that gradients are sparse in their original domain while others find some transformed domains more appropriate. Following the first approach, the scheme proposed in \cite{dgc} sparsified gradients by setting insignificant entries to zero, where insignificant entries were those entries whose values were between top $0.05 \%$ and bottom $0.05 \%$. The authors then used run-length codes to encode the gradients before sending them to the server. In \cite{sbc}, a similar scheme was studied.

The second approach for sparsification aims to find another domain for gradients in which they are more sparse than their original domain \cite{2ndPar2nd}. In \cite{atomo}, gradients were transformed to the Singular Value Decomposition (SVD) domain to discover/exploit more sparsity helping to mitigate the communication overheads. Also, the authors of \cite{atomo} stated that TernGrad \cite{ternGrad} and QSGD \cite{qsgd} were special versions of their proposed scheme under certain circumstances. The authors in \cite{bigMatrix} proved why in general, big matrices are often low-rank.

In the literature, each of the aforementioned techniques, i.e., local updates and gradient compression, has been {\it individually} demonstrated to mitigate the communication overheads in FL, leading to considerable speedups.
Specifically, dynamically determining local update coefficients to balance the communication-computation trade-off has been studied in \cite{adacom, adapjsac, adaptau}, but without considering any compression of gradients. Meanwhile, \textit{static} gradient compression (communication-precision trade-off) has been studied in \cite{2ndPar2nd, atomo}, but without dynamically adjusting local update coefficients. Based on these works, it is expected that conjoining the two techniques would be even more effective in reducing the communication overheads, thereby accelerating FL. To the best of our knowledge, however, studying and balancing those two trade-offs \textit{jointly and dynamically} while considering their impacts on convergence (from both theoretical and empirical perspectives) have not been done in the literature. This unexplored, yet important problem motivated our work. In this paper, we propose such jointly adjusted/balanced scheme, namely Fast FL (FFL), which jointly and dynamically determines the \textit{local update coefficients} and \textit{sparsity budgets} of gradient compression. Our main contributions in this paper are as follows:

\begin{itemize}
	\item As the first work in the literature, we formulate our FL problem to minimize the error of the global model in a given wall-clock time with respect to \textit{both} local update coefficients and sparsity budgets, which are respectively demonstrated to characterize the trade-off between communication and computation and the trade-off between communication and precision.
	
	\item We derive an upper bound of the error of the global model in a given wall-clock time considering the interdependency between the two variables: the local update coefficients and sparsity budgets. 
	
	\item Using the derived error upper bound, we propose an enhanced FL scheme, FFL, which jointly and dynamically determines the two variables to accelerate learning.
	
	\item Our analytical results include almost all of the unbiased compression techniques. This is because our formulation for compression is based on atomic decomposition for sparse representation which is a popular technique for compression from compressed sensing.
	
	\item We demonstrate that FFL consistently achieves higher accuracies faster than similar schemes existing in the literature.
\end{itemize}

The remainder of this paper is organized as follows: in Section \ref{iamsectwo}, the fundamental mechanism of FL is presented and our proposed scheme is described at a high level. In Section \ref{iamsecthree}, our problem is mathematically formulated, followed by derivation of an error upper bound of the learning error. The FFL scheme is proposed in Section \ref{iamsecfour}. Experiment results are provided in Section \ref{iamsecfive} and this paper is concluded in Section \ref{iamsecsix}.

\textit{Notations:} All vectors are column vectors and denoted by bold font small letters (e.g., $\boldsymbol{x}$), while scalars are denoted by normal font small letters (e.g., $y$). Matrices are denoted by bold font capital letters (e.g., $\boldsymbol{A}$). Also, $\boldsymbol{x}^T$ denotes the transpose of $\boldsymbol{x}$. We use “$:=$” to denote “is defined to be equal to.” We also use $\| \cdot \|_1$ and $\| \cdot \|$ to denote the $L_1$ and $L_2$ norms, respectively. For a set $\mathcal{S}$, $|\mathcal{S}|$ denotes its cardinality. We use $\nabla F(\boldsymbol{x})$ to represent the gradient of $F(\boldsymbol{x})$. Expectation with respect to random variable $X$ is denoted by $\mathbb{E}_{X}\left[ \cdot \right]$.

\section{FL Mechanism and the Proposed Scheme} \label{iamsectwo}
In this section, we first present the fundamental learning mechanism of FL and then high level descriptions of our proposed scheme.
\subsection{FL Mechanism}
Consider an $N$-sample-size dataset defined by $ \mathcal{S} = \{  (\boldsymbol{x}_1,y_1) , (\boldsymbol{x}_2,y_2 ), \cdots,  (\boldsymbol{x}_N,y_N) \}$, where the pair of $ (\boldsymbol{x}_q,y_q) $ includes the $q$th data sample $\boldsymbol{x}_q$ and its corresponding label $y_q$. In the FL setting, the dataset $\mathcal{S}$ is distributed among $M$ distinct workers, each of which holds the local dataset $\mathcal{S}_j \subset \mathcal{S}$ for $j=1,2, \cdots, M $, where $ \cup_{j=1}^M \mathcal{S}_j = \mathcal{S}$ and $\mathcal{S}_{i^{\prime}} \cap \mathcal{S}_{j^{\prime}} = \emptyset $, for $i^{\prime} \neq j^{\prime}$. In the $k$th round, the global loss function $F(\boldsymbol{w}_k; \mathcal{S})$ on all of the distributed datasets is given by
\begin{equation}
F(\boldsymbol{w}_{k}; \mathcal{S}) := \frac{1}{\left|\mathcal{S}\right|} \sum_{(\boldsymbol{x}_q , y_q ) \in \mathcal{S}} f (\boldsymbol{w}_{k}; \boldsymbol{x}_q , y_q )
\label{eq:globlossfun}
\end{equation}
for $k=0,1,2,\cdots, K$, where $\boldsymbol{w}_k \in \mathbb{R}^d$ is the global weight vector in the $k$th round for the global model, $d$ denotes the dimension of the weight vector, and $K$ is the last round. Also, $f (\boldsymbol{w}_{k}; \boldsymbol{x}_q , y_q) \in \mathbb{R}$ stands for the value of loss function for the $q$th data sample. Because $\mathcal{S}_j$'s are distributed over multiple distinct workers, it is not possible for the server to directly minimize the global loss function in (\ref{eq:globlossfun}). Instead, in FL, each worker minimizes its own local loss function $F(\boldsymbol{w}_{k}^{j}; \mathcal{S}_j)$, which is again defined by (\ref{eq:globlossfun}) but with local dataset $\mathcal{S}_j$ and local weight vector $\boldsymbol{w}_{k}^{j} \in \mathbb{R}^d$ for its local model.

The FL learning process proceeds iteratively as follows: at the initialization stage ($k = 0$), all local weight vectors $\boldsymbol{w}_{0}^{j}$'s at different workers are initialized to the same value. At the beginning of each round (for $k > 0$), new values of local weight vectors are computed by performing $\tau_k$ number of consecutive local updates via SGD at the workers. Specifically, at the $\ell$th local update, the local weight vector is updated from $\boldsymbol{w}^{j, \ell-1}_{k}$ to $\boldsymbol{w}^{j, \ell}_{k}$ as follows:
\begin{equation}
\boldsymbol{w}_{k}^{j,\ell}  = \boldsymbol{w}_{k}^{j,\ell-1} -\eta  \boldsymbol{g}(\boldsymbol{w}^{j,\ell-1}_{k}; \xi_{j})
\label{eq:local_update_eq}
\end{equation}
for $j=1,2, \cdots, M$,  $k=0,1,2,\cdots, K$, and $\ell = 1, 2, \cdots, \tau_k$, where $\eta$ is the learning rate and $\xi_{j} \subset \mathcal{S}_j$ is a randomly selected mini-batch with replacement at the $j$th worker. Also, $\boldsymbol{g}(\boldsymbol{w}_{k}^{j, \ell}; \xi_{j}) = \nabla F(\boldsymbol{w}_{k}^{j, \ell}; \xi_{j})$ is the gradient of $F(\boldsymbol{w}_{k}^{j, \ell}; \xi_{j})$ at the $\ell$th local update calculated with weight vector $\boldsymbol{w}_{k}^{j, \ell}$, on the mini-batch $\xi_{j}$. The initial local weight vector $\boldsymbol{w}_{k}^{j,\ell=0}$ is the global weight vector transmitted from the server in the end of the previous round ($k-1$). %Note that since the mini-batch at each worker is a randomly selected mini-batch with replacement, it has the same sample space for all $k$ and $\ell$, i.e., the probability distribution of the mini-batch $\xi_{j}$ selected at the $j$th worker is independent of $\ell$ and $k$. 
As soon as the $j$th worker performs $\tau_k$ number of local updates according to (\ref{eq:local_update_eq}), the aggregated gradient $\boldsymbol{g} (\boldsymbol{w} _k^j)$ in the $k$th round is determined by 
\begin{equation}
\boldsymbol{g}(\boldsymbol{w}_{k}^{j}):= \sum_{\ell=1}^{\tau_k}\boldsymbol{g}(\boldsymbol{w}_{k}^{j, \ell}; \xi_{j})
\label{eq:localup}
\end{equation}
for $j=1,2, \cdots, M,$ and $k=0,1,2,\cdots, K$. The local update coefficient $\tau_k$ determines the computation (due to local update) to communication ratio in the $k$th round and characterizes the communication-computation trade-off that will be discussed later.

\begin{figure}[t]
	\centering
	\includegraphics[scale=0.6]{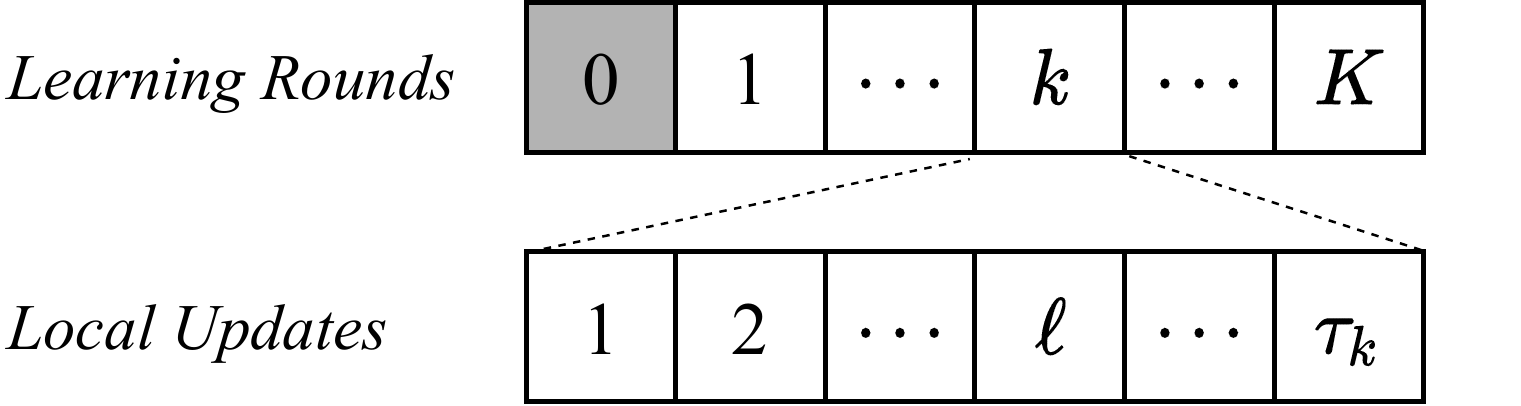}
	\caption{Two time scales of FL. Each round comprises $\tau_k$ number of local updates. The FL system is initialized at $k=0$. }
	\label{fig:periodroundlocalold}
\end{figure}

\begin{figure}[t!]
	\centering
	\includegraphics[scale=0.74]{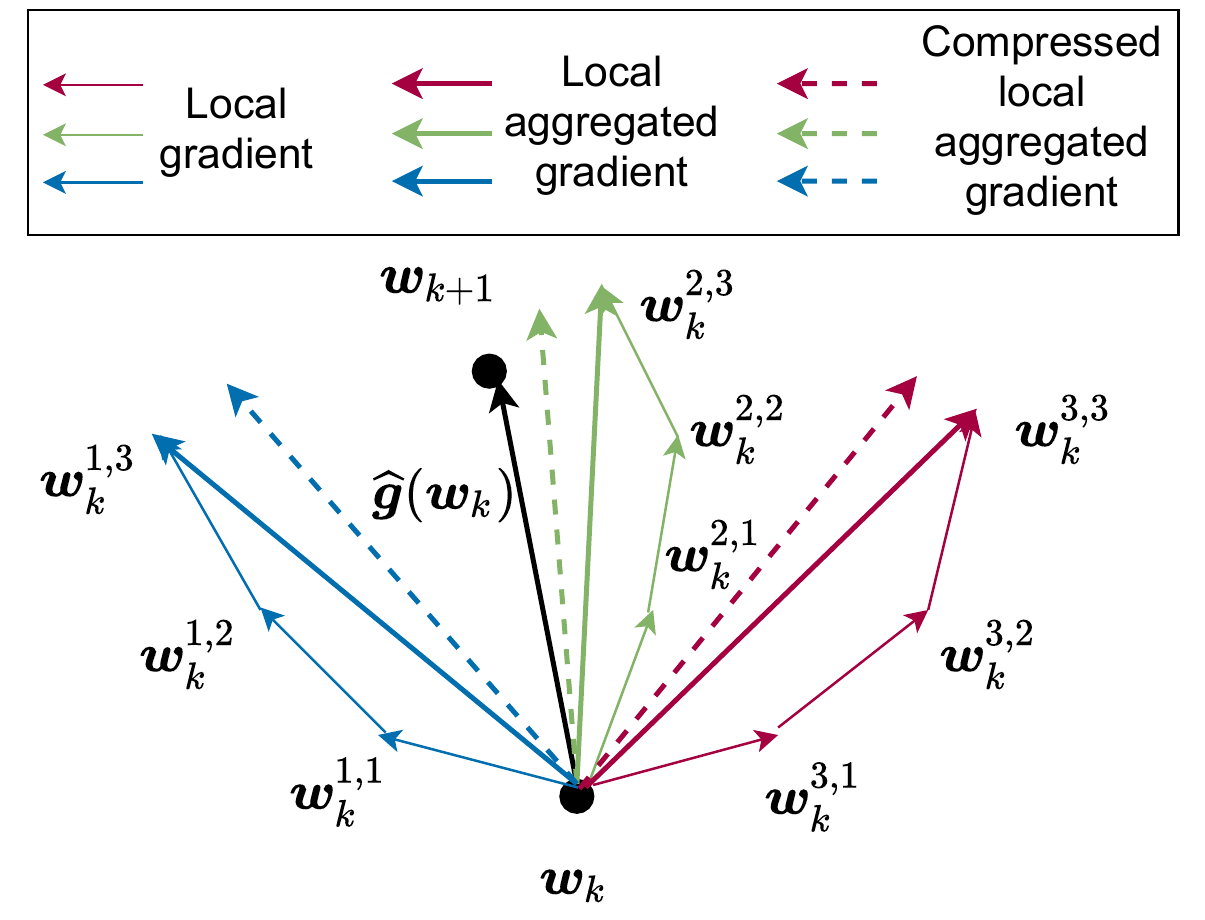}
	\caption{In each round, workers locally update their weights multiple times based on their private datasets. Then they compress the \textit{local aggregated gradient} and send them to the server who averages the received gradients and applies them to the global weight. It is clear that $\tau_k = 3$ and we have $3$ workers.}
	\label{fig:gist}
\end{figure}

\begin{table}[t!]
	\caption{Main Notations.}
	\centering
	\begin{tabular}{|c | l |} 
		\hline
		Notation & Description  \\ [0.5ex] 
		\hline
		$\tau_k$ & Local update coefficient of the $k$th round  \\
		$s_k$ & Sparsity budget of the $k$th round \\ 
		$\boldsymbol{w}_{k}$ & Global weights of the $k$th round \\
		$\boldsymbol{w}_{k}^{j}$ & Weights of the $j$th worker ($k$th round) \\
		$\boldsymbol{w}_{k}^{j, \ell}$ & The $\ell$th local weights of the $j$th worker \\
		$\boldsymbol{g} (\boldsymbol{w}_{k})$ & The global gradient of the $k$th round \\
		$\boldsymbol{g}(\boldsymbol{w}_{k}^{j})$ & Local aggregated gradient of the $j$th worker  \\
		$\boldsymbol{g}(\boldsymbol{w}_{k}^{j, \ell}; \xi_{j})$ & The $\ell$th local update of the $j$th worker \\
		$\widehat{ \boldsymbol{g} }(\boldsymbol{w}_{k})$ & The global compressed gradient ($k$th round)  \\
		$\widehat{ \boldsymbol{g} }(\boldsymbol{w}_{k}^{j})$ & The compressed local gradient ($j$th worker) \\ [1ex] 
		\hline
	\end{tabular}
	\label{tab:tab1}
\end{table}

After calculating the aggregated gradient by (\ref{eq:localup}), all $M$ workers \textit{compress} their locally aggregated gradients $\boldsymbol{g}(\boldsymbol{w}^j_k)$ to $\widehat{ \boldsymbol{g}}(\boldsymbol{w}^j_k)$ with a given \textit{sparsity budget} of gradient compression, $s_k$ (will be defined later), and send them to the server\footnote{In the literature, two different approaches have been used for the global aggregation at the server: gradient-averaging and weight-averaging. The former requires each worker to send the computed gradients to the server, whereas in the latter weights themselves are transmitted. In this paper, we use gradient-averaging because gradients are far more sparse than weights in almost any domain. This helps to reduce the communication overhead, which is the main challenge of FL.}\footnote{If weight-averaging \textit{were} adopted, the workers \textit{would} transmit the latest weights. In our paper, however, the workers transmit the aggregate of local gradients as in (\ref{eq:localup}).}. Note that here the second trade-off, communication-precision trade-off emerges by introducing gradient compression. Then, at the server, the compressed global gradient vector, denoted by $\widehat{ \boldsymbol{g} }(\boldsymbol{w}_k)$, is obtained by averaging all received gradient vectors from the workers as follows:
\begin{equation}
\widehat{ \boldsymbol{g} }(\boldsymbol{w}_{k}) = \frac{1}{M}\sum_{j=1}^{M}\widehat{ \boldsymbol{g} }(\boldsymbol{w}_{k}^{j}), \quad k=0,1,2,\cdots, K.
\label{eq:globalave}
\end{equation}
Using the average of gradient vectors, the global weight vector $\boldsymbol{w}_{k}$ is updated by SGD (or its variants) as follows: 
\begin{equation}
\boldsymbol{w}_{k+1}=\boldsymbol{w}_{k} - \eta  \widehat{ \boldsymbol{g}} (\boldsymbol{w}_{k}), \quad k=0,1,2,\cdots, K. 
\label{eq:globalup}
\end{equation}
The server then broadcasts the updated global weight vector $\boldsymbol{w}_{k+1}$ to synchronize all workers. This is the process of a single round involving $M$ workers, each of which individually performing $\tau_k$ local updates with gradient compression. Fig. \ref{fig:periodroundlocalold} shows the time scales of the rounds and local updates while Fig. \ref{fig:gist} portraits local updates, aggregated gradients, compressed aggregated gradients, and the global update. Table \ref{tab:tab1} lists/describes our main notations. In this section, we have introduced two inherent trade-offs to be balanced for achieving fast FL, and in the next section, we will discuss the trade-offs.

\subsection{Proposed Scheme for FL: Adjusting Two Key Variables $\{ \tau_k \}$ and $\{ s_k \}$} \label{secpropapp}
Achieving ``fast FL'' is the main concern of our proposed scheme. However, achieving ``fast FL'' and ``communication-efficient FL'' should not be conflated: although these two objectives overlap/correlate in practice, they are still different at a conceptual/theoretical level. Achieving fast FL requires communication-efficiency; but, it is not limited to it. For achieving fast FL, every time-consumer should be taken into account: (i) downlink communication time from the server to the workers, (ii) time for local training of neural networks (i.e., computation time) at the workers, and (iii) uplink communication time from workers to the server. In a special case, when the communication data rate is low, the communication time becomes the dominant time-consumer---and the bottleneck. In such a case, ``achieving fast FL'' \textit{reduces} to ``achieving communication-efficient FL.''

The problem of achieving ``fast FL'' can be framed as to jointly and dynamically find balances considering the two inherent trade-offs: (i) communication-computation and (ii) communication-precision. The first trade-off is characterized by the local update coefficient $\tau_k$, for which striking a balance matters because one extreme (high communication and low computation) slows the convergence whereas the other extreme compromises the accuracy \cite{adacom}; either way, the learning is decelerated. The second trade-off is characterized by the sparsity budget of gradient compression where seeking a balance is imperative since high compression/imprecision (one extreme) stalls the learning process while no compression incurs heavy/unnecessary communication overheads. Having that framing of the problem in mind, (in Section \ref{iamsecthree}) we will mathematically formulate our problem to minimize the learning error in a given wall-clock time with respect to local update coefficient and sparsity budget of gradient compression. 

Before delving into formulating our problem, we define two key variables of the proposed scheme: the first one is the local update coefficient $\tau_k$ which was used in (\ref{eq:localup}). As we have mentioned, in the proposed scheme, instead of fixing the local update coefficient $\tau_k$ over different rounds, we dynamically adjust it in the course of the entire learning process to facilitate the convergence while minimizing the dispensable gradient transmissions.

The second key variable is called the sparsity budget. Assume $\boldsymbol{A} \in \mathbb{R}^{m \times n}$ is a big matrix, which needs to be transmitted. To minimize the communication overheads, instead of sending the exact matrix $\boldsymbol{A}$ as is, we send its approximation $\widehat{\boldsymbol{A}}$ that is constructed only by $s_k$ number of selected \textit{basis components}, which are obtained by the \textit{compressor}. In the real-world applications, for most domains, due to sparsity, many (or even most) of the basis components of big matrices are often small and negligible \cite{bigMatrix}, and in this case, $\widehat{\boldsymbol{A}}$ can precisely approximate $\boldsymbol{A}$ by judiciously selecting $s_k$ basis components. In the context of the proposed FL, sending an approximation of matrix $\boldsymbol{A}$ means sending an approximation of vector $\boldsymbol{g} (\boldsymbol{w}_{k}^{j})$ of all users.

After all, in the proposed scheme, both variables $\{ \tau_k \}$ and $\{ s_k \}$ will be jointly and dynamically optimized over different rounds, while considering their interdependency. Note that adjusting the local update coefficients $ \{\tau_k \}$ was studied in \cite{adacom}, but without considering compression of $\boldsymbol{g} (\boldsymbol{w}_{k}^{j})$, not to mention the \textit{dynamic} compression (characterized by $\{ s_k \}$). Meanwhile, it was studied that compression could reduce the communication overheads in FL \cite{atomo, qsgd, one, ternGrad}. However, gradient compression was not dynamically controlled; that is, $\{ s_k \}$ were assumed to be fixed to a pre-determined $s$. Furthermore, in \cite{atomo, qsgd, one, ternGrad}, the local updates were not adopted; that is, $\{ \tau_k \}$ were assumed to be all fixed to one. To the best of our knowledge, jointly adjusting $\{ \tau_k \}$ and $\{ s_k \}$ has not been studied in the literature, which motivated our work.

\section{Problem Formulation and the Error Upper Bound Analysis} \label{iamsecthree}
In this section, we first mathematically formulate our fast FL problem and then derive a mathematical expression for the error upper bound of the FL loss function in a given wall-clock time, which will be needed to develop the proposed scheme in the next section.

\subsection{Problem Formulation}
In the proposed scheme, both $\{ \tau_k \}$ and $\{ s_k \} $ are jointly and dynamically optimized. Mathematically, the problem is to determine the optimal values of both $\{ \tau_k \}$ and $\{ s_k \} $ in different rounds so that the error in a given wall-clock time is minimized. This is formulated as follows:
\begin{align}
& \min _{ \{\tau_k\}, \{s_k\} } \mathbb{E}_{\{ \xi_{j} \}} \left[\min _{k \in \{ 1,2, \cdots ,K \}} F (\boldsymbol{w}_{k};\mathcal{S}) \right] \notag \\
& \quad \textrm{s.t.} \quad \sum_{k=1}^{K} ( D_k + Y_k)  = T
\label{eq:start}
\end{align}
where $\tau_k \in \{1, 2, \cdots, \tau_{ub} \}$ and $s_k \in [1, s_{ub}]$ represent the local update coefficient and the sparsity budget of the $k$th round, respectively. Also, $F(\boldsymbol{w}_{k};\mathcal{S})$ denotes the global loss function defined in (\ref{eq:globlossfun}) with the global weight vector $\boldsymbol{w}_{k}$ in the $k$th round and dataset $\mathcal{S}$. Communication and computation times in the $k$th round are denoted by $D_k$ and $Y_k$. The given wall-clock time within which the global loss function is to be minimized is denoted by $T$. This time constraint helps to avoid trivial solutions: (i) a solution with too large values of local update coefficients that would be time-consuming and/or (ii) a solution with its sparsity budgets set to the maximum value that extremely/unnecessarily aggravates communication overheads and consumes time. For solving the optimization problem defined in (\ref{eq:start}), we need the expression of learning error in terms of both $\{ \tau_k \}_{k=1}^{K}$ and $\{ s_k \}_{k=1}^{K}$ after $K$ rounds in a given wall-clock time $T$. It is generally impossible to find such an exact analytical expression \cite{adacom, adapjsac, adaptau}. It is even difficult to find a upper bound of the learning error in terms of both $\{ \tau_k \}_{k=1}^{K}$ and $\{ s_k \}_{k=1}^{K}$ after $K$ rounds. In this paper, to make the analysis tractable, we will derive an upper bound of the learning error (i.e., the cost function of (\ref{eq:start})), in terms of both $\tau_k$ and $s_k$  after each round in a given wall-clock time. This provides the mathematical expression that will be needed to optimize the values of $ \tau_k $ and $ s_k$.

\subsection{Error Upper Bound Analysis}
In this subsection, we first derive an upper bound for the learning error after a given wall-clock time as a function of $ \tau_k $ {\it without} considering the gradient compression that is characterized by $s_k$. After then, the effects of gradient compression will be reflected on the upper bound. Conjoining local update and gradient compression presents four theoretical challenges, which are addressed in our paper: (i) gradient compression must be unbiased to ensure convergence of the learning, which is the same condition that we have for SGD to be an unbiased estimation of Full-batch Gradient Descent (FGD). (ii) The variance between compressed gradient and uncompressed gradient must be made as small as possible because this way it has been shown that the learning is accelerated \cite{atomo}. (iii) The communication time has to be written in terms of sparsity budget as it becomes dependent on the sparsity budget of gradient compression. (iv) Gradient compression induces a variance\footnote{The gradient compressor we are using in this paper is a probabilistic compressor whose mean is the same as the uncompressed gradient (unbiased). However, the gradient compressor induces a variance from the uncompressed gradient. If we did not use gradient compression before transmission, there would be no such a variance.} from uncompressed gradient (pure SGD) that needs to be incorporated into the variance of SGD. Note that SGD itself has a variance from FGD. We need to merge these two variances.

From \cite{cooperativesgd, adacom}, we present Theorem 1 (with some adjustment of notation fitting our scheme), which holds if the following three assumptions hold: (i) the global loss function $F(\boldsymbol{x}; \mathcal{S})$ is differentiable, and Lipschitz smooth, which means $\|\nabla F(\boldsymbol{x}; \mathcal{S})-\nabla F(\boldsymbol{y}; \mathcal{S} )\| \leq L\|\boldsymbol{x}-\boldsymbol{y}\|$, with a lower bound $F_{\rm inf}$; (ii) the SGD is an unbiased estimator of the FGD $\mathbb{E}_{\{ \xi_{j} \} }[\boldsymbol{g}(\boldsymbol{w}_{k})]=\nabla F(\boldsymbol{w}_{k}; \mathcal{S})$ for all $k$; and (iii) the variance of the calculated global mini-batch gradient is bounded as $\mathbb{E}_{ \{ \xi_{j} \} } [ \|\boldsymbol{g}(\boldsymbol{w}_{k})-\nabla F(\boldsymbol{w}_{k}; \mathcal{S})\|^{2} ] \leq \beta\|\nabla F(\boldsymbol{w}_{k}; \mathcal{S})\|^{2}+\sigma$, for all $k$, where $\beta$ and $\sigma$ are non-negative constants and inversely proportional to the mini-batch size. In this inequality, $\sigma$ represents the variance between the SGD, $\boldsymbol{g}(\boldsymbol{w}_{k})$, and the FGD, $\nabla F(\boldsymbol{w}_{k}; \mathcal{S})$.

\noindent  \textbf{Theorem 1} (Error Upper Bound \textit{without} Gradient Compression \cite{adacom})\textbf{:}
\textit{Let $Y_k$ and $D_k$ denote the computation and communication times at the $k$th round, respectively. If the learning rate satisfies $\eta L+\eta^{2} L^{2} \tau_k (\tau_k -1) \leq 1$, $Y_k$ and $D_k$ are constant in the $k$th round, and the weight vectors of all workers are initialized at the same point $\boldsymbol{w}_{k}$, then after $T_k$ wall-clock time, the expression in (\ref{eq:start}) over the $k$th round will be bounded by:}
\begin{align}
\frac{2\left[F\left(\boldsymbol{w}_{k}\right)-F_{\rm inf}\right]}{\eta T_k}\left(Y_k+\frac{D_k}{\tau_k}\right) +\frac{\eta L \sigma}{M}+  \eta^{2} L^{2} \sigma(\tau_k -1) \quad \quad \quad \quad \label{eq:theo1}
\end{align}

\noindent \textit{where $L$ is the Lipschitz constant of the loss function.}  

\noindent \textit{Proof: See Appendix of \textnormal{\cite{adacom}}.}  \hfill $\square$

\noindent Theorem 1 specifies the dynamics of our first trade-off, communication-computation trade-off. As we mentioned, this trade-off is characterized by $\tau_k$ which determines the computation to communication ratio. In the error upper bound of (\ref{eq:theo1}), the variable $\tau_k$ is both in numerator and denominator, which indicates that both too small and large values of $\tau_k$ can contribute in increasing the error upper bound. Therefore, seeking a balance is necessary. Also, due to presence of the loss value $F\left(\boldsymbol{w}_{k}\right)$ in the error upper bound and noting that the value of loss is time-varying, the trade-off must be balanced dynamically over the course of learning.

Theorem 1 would be enough if the notion of local update were solely used \textit{without} considering the gradient compression---communication-precision trade-off. In the proposed scheme, however, we also adopt gradient compression, which complicates the error upper bound analysis. Specifically, incorporating compression to the \textit{aggregated local gradient} of the local updates before each transmission complicates the error upper bound given by Theorem 1 in two ways: (i) the communication time $D_k$ becomes dependent on sparsity budget $s_k$. (ii) Compression introduces extra imprecisions (extra terms) to the variance $\sigma$ in (\ref{eq:theo1}): the variance becomes dependent on the sparsity budget $s_k$.

As discussed in Section \ref{secpropapp}, for compression, the matrix to be sent is written as a weighted sum of multiple \textit{atom matrices} (basis components), where a \textit{compressor}\footnote{The compressor can be any compressor that is based on the atomic decomposition for sparse representation in compressed sensing.} is used to extract the atom matrices. Due to sparsity \cite{atomo}, some atom matrices have more contribution than others in precisely approximating the original matrix. Therefore, the problem is to perform an unbiased selection of atom matrices so that the variance\footnote{Note that this variance is distinct from the variance (denoted by $\sigma$) mentioned earlier. This variance characterizes the variance of compression process by the estimator whereas the variance denoted by $\sigma$ characterized the variance of mini-batch SGD. Later, in Theorem 3, we will merge these two variances.} is minimized. Accordingly, we write the $j$th worker's gradient $\boldsymbol{g}(\boldsymbol{w}_{k}^{j})$ as follows:
\begin{equation}
\boldsymbol{g}(\boldsymbol{w}_{k}^{j})=\sum_{i=1}^{B} \lambda^{i}(\boldsymbol{w}_{k}^{j}) \boldsymbol{a}^{i}(\boldsymbol{w}_{k}^{j})
\end{equation}
where $\boldsymbol{a}^{i}(\boldsymbol{w}_{k}^{j}) \in \mathbb{R}^d$ is the $i$th atom, $\lambda^{i}(\boldsymbol{w}_{k}^{j})$ is its corresponding coefficient, and $B$ is the number of atom matrices that are summed. Note that, for notational simplicity, we assume that atoms matrices are flattened; that is why $\boldsymbol{a}^{i}(\boldsymbol{w}_{k}^{j}) \in \mathbb{R}^d$. Note that our formulation for compression relies on writing a matrix as a combination of atom matrices. In \textit{compressed sensing}, this is called \textit{atomic decomposition for sparse representation}. The formulation of compression via atomic decomposition makes our work widely inclusive of almost all of the unbiased compression techniques: specifically, TernGrad \cite{ternGrad} and QSGD \cite{qsgd}, two important quantization schemes, are special cases of this formulation\footnote{This has been shown in \cite{atomo}.}. Meanwhile, sparsification techniques such as spectral-ATOMO \cite{atomo} and element-wise sparsification (e.g., top-$k$ in DGC \cite{dgc}) also comply with this formulation. Now, the problem is how and which coefficients, $\lambda^{i}(\boldsymbol{w}_{k}^{j})$'s, should be selected. As the first requirement, we are interested in an ``estimator'' that is unbiased. The following estimator qualifies this requirement of unbiasedness (adopted from \cite{atomo} with some adaptation):
\begin{equation}
\widehat{\boldsymbol{g}}(\boldsymbol{w}_{k}^{j})=\sum_{i=1}^{B} \frac{\lambda^{i}(\boldsymbol{w}_{k}^{j}) e^{i}(\boldsymbol{w}_{k}^{j})}{p^{i}(\boldsymbol{w}_{k}^{j})} \boldsymbol{a}^{i}(\boldsymbol{w}_{k}^{j})
\label{eq:estimator}
\end{equation}
where $e^{i}(\boldsymbol{w}_{k}^{j}) \sim \operatorname{Bernoulli} \left(p^{i}(\boldsymbol{w}_{k}^{j})\right)$ and $e^{i}(\boldsymbol{w}_{k}^{j})$'s are independent, for $0 < p^{i}(\boldsymbol{w}_{k}^{j}) \leq 1$. Also, $p^{i}(\boldsymbol{w}_{k}^{j})$ denotes the probability characterizing the Bernoulli distribution, which will be optimized later. We derive two key properties for the estimator in (\ref{eq:estimator}): (i) the estimator of (\ref{eq:estimator}) is unbiased as it is presented in Lemma 1. Unbiasedness is necessary for guaranteeing theoretical convergence as it was a requirement of Theorem 1. (ii) The variance of the estimator in (\ref{eq:estimator}) is derived in Lemma 2.

\noindent \textbf{Lemma 1} (Unbiased Estimator)\textbf{:}
\textit{The estimator given in \textnormal{(\ref{eq:estimator})} is unbiased: $\mathbb{E}_{ \{ e^{i}(\boldsymbol{w}_{k}^{j}) \} }[\widehat{\boldsymbol{g}}(\boldsymbol{w}_{k}^{j})] = \boldsymbol{g}(\boldsymbol{w}_{k}^{j})$.}

\noindent \textit{Proof: The proof is straightforward via the definition of expectation.}  \hfill $\square$

\noindent \textbf{Lemma 2} (Variance of Estimator)\textbf{:}
\textit{The variance of estimator in \textnormal{(\ref{eq:estimator})} is given by $\mathbb{E}_{ \{ e^{i}(\boldsymbol{w}_{k}^{j}) \} } [\| \widehat{\boldsymbol{g}}(\boldsymbol{w}_{k}^{j})-\boldsymbol{g}(\boldsymbol{w}_{k}^{j})\|^2] = \sum_{i=1}^{B} \lambda^{i}(\boldsymbol{w}_{k}^{j})^2  \left(\frac{ 1}{p^{i}(\boldsymbol{w}_{k}^{j})} -1 \right)$.}

\noindent \textit{Proof: See Appendix A.}  \hfill $\square$

Having derived the variance of the estimator, we can formulate an optimization problem to minimize the variance. Note that making the variance of the estimator as small as possible is important since it is known that the smaller the variance, the closer the estimated/compressed gradient $\widehat{\boldsymbol{g}}(\boldsymbol{w}_{k}^{j})$ is to the uncompressed one $\boldsymbol{g}(\boldsymbol{w}_{k}^{j})$, which leads to faster convergence of training \cite{atomo}. In the optimization problem to be formulated, the variables to be determined are $p^{i}(\boldsymbol{w}_{k}^{j})$'s. Thus, the optimization problem is given by
\begin{align}
\min \sum_{i=1}^{B} \frac{\lambda^{i}(\boldsymbol{w}_{k}^{j})^{2}}{p^{i}(\boldsymbol{w}_{k}^{j})} \ \text {s.t.} \ 0<p^{i}(\boldsymbol{w}_{k}^{j}) \leq 1 \  \text {and}  \ \sum_{i=1}^{B} p^{i}(\boldsymbol{w}_{k}^{j})=s_k.
\label{formu:sixth}
\end{align}
Under the assumption of $s_k$-balancedness in the following definition, the solution to the optimization problem in (\ref{formu:sixth}) is derived in Theorem 2.

\noindent \textbf{Definition 1} ($s_k$-balancedness)\textbf{:} \textit{An atomic decomposition $\boldsymbol{g}(\boldsymbol{w}_{k}^{j})=\sum_{i=1}^{B} \lambda^{i}(\boldsymbol{w}_{k}^{j}) \boldsymbol{a}^{i}(\boldsymbol{w}_{k}^{j})$ is $s_k$-unbalanced at the $i$th entry if $ \lambda^{i}(\boldsymbol{w}_{k}^{j}) s_k > \| \lambda(\boldsymbol{w}_{k}^{j}) \|_1 $. If at no entry $\boldsymbol{g}(\boldsymbol{w}_{k}^{j})$ is $s_k$-unbalanced, then we call it $s_k$-balanced.}

\noindent \textbf{Theorem 2} (Solution to the Optimization Problem in (\ref{formu:sixth}))\textbf{:}
\textit{If $\boldsymbol{g}(\boldsymbol{w}_{k}^{j})$ is $s_k$-balanced, the solution to the optimization problem in (\ref{formu:sixth}) is given by}
\begin{equation}
p^{i}(\boldsymbol{w}_{k}^{j})= \frac{\lambda^{i}(\boldsymbol{w}_{k}^{j})s_k}{\| \lambda (\boldsymbol{w}_{k}^{j}) \|_1}.
\end{equation}
\noindent \textit{Proof: This can be proven via Lagrangian multiplier.}  \hfill $\square$

After introducing the estimator in (\ref{eq:estimator}) and deriving the optimal probabilities $p^{i}(\boldsymbol{w}_{k}^{j})$'s minimizing the estimator's variance in Theorem 2, we can now establish the effects of compression on (i) communication time $D_k$ and (ii) variance $\sigma$. For communication time $D_k$, assuming that $\boldsymbol{A} \in \mathbb{R}^{m \times n}$ is a matrix of gradients to be sent, instead of $\boldsymbol{A}$, one can send $\widehat{\boldsymbol{A}}$ with sparsity budget $s_k$, i.e., only $s_k$ number of \textit{atoms} are sent. As a result, we have communication time $D_k$ as a function of sparsity budget $s_k$ given by $D_k=\alpha s_k$, where $\alpha$ is the communication time per atom. For variance $\sigma$, based on Theorem 2, which gave the solution to the optimization problem in (\ref{formu:sixth}), we derive the variance of \textit{compressed aggregated gradient} in the following theorem.

\noindent \textbf{Theorem 3} (Variance of the Compressed Gradient SGD)\textbf{:}
\textit{The variance of the compressed gradient SGD is bounded as follows:} 
\begin{align}
\mathbb{E}_{ \{ \xi_{j} \} , \{ e^{i}(\boldsymbol{w}^j_{k})  \} } \Big[
\|\widehat{\boldsymbol{g}}(\boldsymbol{w}_{k})-\nabla F(\boldsymbol{w}_{k})\|^{2} \Big] \notag \\
 \leq \beta\|\nabla F(\boldsymbol{w}_{k})\|^{2}+ \frac{\sigma_1}{s_k}  +\sigma_2
\label{eq:theoo3}
\end{align}
\textit{where $\beta$, $\sigma_1$, and $\sigma_2$ are non-negative constants and inversely proportional to the mini-batch size.}

\noindent \textit{Proof: See Appendix B.}  \hfill $\square$

Now we can re-write the third assumption of Theorem 1 for the case where the gradient compression is performed, as follows: the upper bound on the variance of compressed SGD evaluated on a mini-batch from $\mathcal{S}_j$ is given by (\ref{eq:theoo3}). Note that, in the new assumption, $\sigma$ is replaced by $\sigma_1 /s_k+\sigma_2$, which does not affect the proof of Theorem 1 given in \cite{adacom}. Therefore, with a change of variable $\sigma = \sigma_1 /s_k +\sigma_2$, the same proof can be exploited. As a result of considering the effects of compression on communication ($D_k=\alpha s_k$) and variance ($\sigma = \sigma_1 /s_k+\sigma_2$), the following theorem which is an extension of Theorem 1 can be proved.

\noindent \textbf{Theorem 4} (Error Upper Bound \textit{with} Gradient Compression)\textbf{:}
\textit{If the learning rate satisfies $\eta L+\eta^{2} L^{2} \tau_k (\tau_k -1) \leq 1$, $Y_k$ and $D_k=\alpha s_k$ are constants in the $k$th round, and the weight vectors of all workers are initialized at the same point $\boldsymbol{w}_{k}$, then after $T_k$ wall-clock time, the expression in (\ref{eq:start}) over the $k$th round will be bounded by:}
\begin{align}
	\psi_k(\tau_k, s_k) = \frac{2\left[F\left(\boldsymbol{w}_{k}\right)-F_{\rm inf}\right]}{\eta T_k}\left(Y_k +\frac{\alpha s_k}{\tau_k}\right) \notag \\
	+  \frac{\eta L (\frac{\sigma_{1}}{s_k}+ \sigma_{2})}{M} +  \eta^{2} L^{2} (\frac{\sigma_{1}}{s_k} + \sigma_{2})(\tau_k -1) \label{eq:keyeqpsi}
\end{align}
\noindent \textit{where $\psi_k(\tau_k, s_k)$ is the upper bound of the error in the $k$th round.}

\noindent \textit{Proof: Similar to the proof of Theorem 1 \textnormal{\cite{adacom}} except that $\sigma$ is replaced by $\sigma_1 /s_k +\sigma_2$, and $D_k=\alpha s_k$.}  \hfill $\square$

This upper bound $\psi_k(\tau_k, s_k)$ in (\ref{eq:keyeqpsi}) specifies the dynamics of our \textit{both} trade-offs: trade-offs between communication and computation/precision characterized by $\tau_k $ and $ s_k $, respectively. Also, $\psi_k(\tau_k, s_k)$ gives us insight into how to balance these trade-offs dynamically over the course of learning. One extreme is when $\tau_k$ is set to its minimum value $\tau_k=1$, where after a single local update (computation), the gradients are communicated. The other extreme is when $\tau_k$ is set to a very large value $\tau_k \gg 1$ (which corresponds to too many computations/local updates). The first term in (\ref{eq:keyeqpsi}) shows the positive effects of employing local updates on the error upper bound $\psi_k(\tau_k, s_k)$. Specifically, the first term implies that the larger the value of $\tau_k$, the smaller the $\psi_k(\tau_k, s_k)$ (as $\tau_k$ is in the denominator). But, larger $\tau_k$ comes with a repercussion that manifests itself in the third term which shows that employing local updates causes an error as a result of a growing discrepancy among local models due to less often communication (synchronization). Therefore, the aim is to strike a balance between these two extremes: too many communications when $\tau_k=1$ and too many computations for $\tau_k \gg 1$.

For $s_k$, one extreme is to set $s_k$ to its minimum value $s_k=1$, where the communication overhead is at its minimum (due to high compression) and consequently the precision of the communicated gradients is the minimum. The other extreme is when $s_k \gg 1$, i.e., communicating precise gradients---high communication overheads. The first term in (\ref{eq:keyeqpsi}) entails reducing $s_k$ (which is in the numerator), thereby mitigating the communication overheads, whereas the second and third terms require the value of $s_k$ to be large (which is in the denominator) so that the introduced imprecision as a result of compression is minimized. Neither $s_k =1$ nor $s_k \gg 1$ is an optimal choice: the former communicates too imprecise gradients resulting in a prolonged convergence, while the latter incurs too much dispensable communication overheads. The goal is to find the optimal balance between these two extremes.

\section{Proposed FL Scheme: FFL} \label{iamsecfour}
We now propose our enhanced FL scheme, namely, FFL, which jointly and dynamically adjusts the values of $\tau_k $ and $s_k $ over different rounds in order to reduce the convergence time. Mathematically speaking, in the $k$th round, FFL minimizes the upper bound of the error in (\ref{eq:keyeqpsi}) as follows: 
\begin{equation}
\tau^{*}_{k}, s^{*}_{k}=\arg \min_{\tau_k, s_k} \ \psi_k(\tau_k, s_k) , \quad k=  1, 2 , \cdots, K \label{eq:optii}
\end{equation}
where $\tau_k \in \{1, 2, \cdots , \tau_{ub} \} $ and $s_k \in [ 1, s_{ub} ]$. We present Theorem 5 that proves the convexity of $\psi_k(\tau_k, s_k)$, and Theorem 6 that gives the optimal and approximate solutions to the problem in (\ref{eq:optii}).

\noindent  \textbf{Theorem 5} (Convexity of $\psi_k(\tau_k, s_k)$)\textbf{:}
\textit{If assumptions (i) $\tau_k \geq 2$, (ii) $\eta^5 \approx 0$, (iii) $(L^4 T_k \sigma_{1} / 2 \alpha (F\left(\boldsymbol{w}_{k}\right)-F_{\rm inf}) s_k^4) < \infty$, and (iv) $2\eta^2 L T_k \sigma_{1} \tau_k \geq \alpha M s_k^2 (F\left(\boldsymbol{w}_{k}\right)-F_{\rm inf})$ hold, then $\psi_k(\tau_k, s_k)$ is convex.}

\noindent \textit{Proof: See Appendix C.} \hfill $\square$

\noindent In the above theorem, assumption (i) excludes only one value of $\tau_{k}$ which is $\tau_{k}=1$. Nonetheless, this does not matter because when $\tau_{k}=1$, the learning has already converged---we will see this in Section \ref{iamsecfive}. Assumption (ii) is reasonable because $0 < \eta \ll 1$. Assumption (iii) always holds in practice. Finally, assumption (iv) excludes only a half-space of the involved parameters.

\begin{figure*}
	\includegraphics[width=\textwidth,height=5cm]{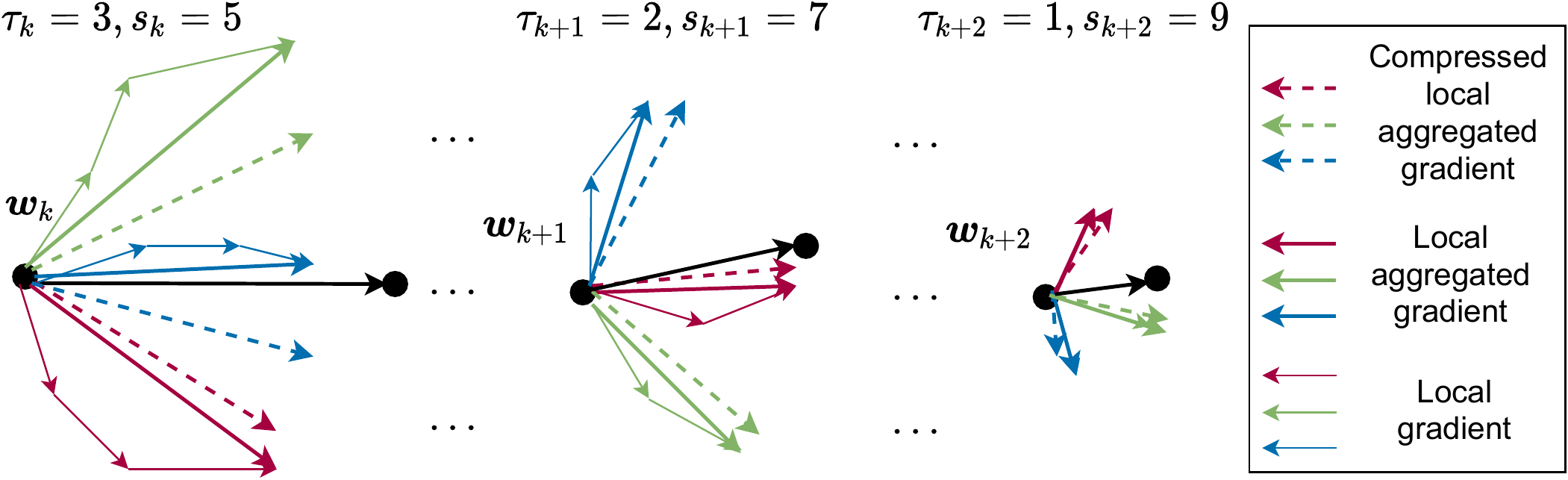}
	\caption{Over the course of learning, the value of $\tau_{k}$ decreases whereas the value of $s_k$ increases.}
	\label{fig:portray}
\end{figure*}

\noindent  \textbf{Theorem 6} (Optimal and Approximate Values of $\tau_{k}$ and $s_k$)\textbf{:}
\textit{The value of upper error bound $\psi_k(\tau_k, s_k)$ is minimized after $T_k$ wall-clock time when the $\tau_{k}$ and $s_k$ are as follows:}
\begin{align}
	\tau_k & = \sqrt{\frac{2 \alpha \left[F\left(\boldsymbol{w}_{k}\right)-F_{\rm inf}\right] s_k^2}{\eta^3 L^2 (\sigma_1 + \sigma_{2}s_k)}} \\ s_k  & = \sqrt{\frac{\sigma_1 \eta^2 L T_k(1 - \eta L (\tau_{k} -1)) \tau_k}{2 \alpha \left[F\left(\boldsymbol{w}_{k}\right)-F_{\rm inf}\right]}}.
\end{align}
\textit{Under assumptions (a) $\sigma_{1} \ll \sigma_{2}s_k$ and (b) $\eta L (\tau_k -1) \ll 1$, the approximate values of $\tau_{k}$ and $s_k$ which involve no hyperparameters are given as follows:}
\begin{align}
	\frac{\tau_{k+1}}{\tau_{k}} & = \sqrt{\frac{F\left(\boldsymbol{w}_{k+1}\right)-F_{\rm inf}}{F\left(\boldsymbol{w}_{k}\right)-F_{\rm inf}}} \sqrt{\frac{\sigma_1 + \sigma_{2}s_k}{\sigma_1 + \sigma_{2}s_{k+1}}} \frac{s_{k+1}}{s_k} \notag \\ & \approx \sqrt{\frac{F\left(\boldsymbol{w}_{k+1}\right)}{F\left(\boldsymbol{w}_{k}\right)}}  \sqrt{\frac{s_{k+1}}{s_k}} \label{eq:appro1} \\
	\frac{s_{k+1}}{s_{k}} & = \sqrt{\frac{F\left(\boldsymbol{w}_{k}\right)-F_{\rm inf}}{F\left(\boldsymbol{w}_{k+1}\right)-F_{\rm inf}}} \sqrt{\frac{1 - \eta L (\tau_{k+1} -1)}{1 - \eta L (\tau_{k} -1)}} \sqrt{\frac{\tau_{k+1}}{\tau_k}} \notag \\ & \approx\sqrt{\frac{F\left(\boldsymbol{w}_{k}\right)}{F\left(\boldsymbol{w}_{k+1}\right)}}  \sqrt{\frac{\tau_{k+1}}{\tau_k}}. \label{eq:appro2}
\end{align}
\noindent \textit{Proof:} \textit{This can be proven by setting partial derivatives of $\psi_k(\tau_k, s_k)$ to zero and applying the assumptions.}
\hfill $\square$

In Theorem 6, assumption (a) is justifiable because $\sigma_2$ is often considerably larger than $\sigma_1$---besides, we know that $s_k \geq 1$. To explain the reason why $\sigma_{2} \gg \sigma_1$, we first answer the question that where do $\sigma_2$ and $\sigma_1$ come from? When using gradient compression in addition to local update SGD for FL, there are two sources of variance with respect to the FGD: (i) calculating the gradients based on mini-batch SGD (instead of FGD) accounts for the first variance, $\sigma_2$, and (ii) performing compression on top of SGD causes the second variance from the FGD, $\sigma_1$. For sparse matrices (e.g., gradients)\footnote{The sparsity of gradients (our supposition) has been observed and exploited in many studies \cite{atomo, dgc, bigMatrix}.}, compression does not distort the matrix considerably because with few \textit{atom} matrices, one can almost precisely represent/reconstruct the original matrix. Therefore, the compression does not have to resort to an aggressively lossy compression. Expectedly, the variance $\sigma_1$, which is introduced as a result of compression is negligible (close to zero) compared to the variance of SGD $\sigma_2$. Meanwhile, the value of $\sigma_{2}$ depends on the mini-batch size. Specifically, we know that the variance of the mini-batch SGD is inversely proportional to the mini-batch size \cite{adacom, cooperativesgd}. Because the mini-batch size of SGD is often significantly smaller than the size of the full-batch, the variance of mini-batch SGD $\sigma_2$ becomes large and therefore it is safe to infer that $\sigma_{2}$ is larger than the variance of compression $\sigma_{1}$. Assumption (b) of Theorem 6 is reasonable as in practice the learning rate is usually small (near $0.01$), $L < 1$, and $(\tau_k -1) < 50$. Also, $F_{\rm inf}$ is usually considered to be zero \cite{atomo, adacom}. Recall that in Theorem 6, assumptions (a) and (b) help to remove the dependency of our optimal solution to hyperparameters. Consequently, Equations (\ref{eq:appro1}) and (\ref{eq:appro2}) help to jointly and dynamically balance trade-offs between communication and computation/precision over the course of learning. The values of $\tau_k$ and $s_k$ in (\ref{eq:appro1}) and (\ref{eq:appro2}) are mutually dependent whose dependence can be decoupled by substituting (\ref{eq:appro1}) in (\ref{eq:appro2}). The results are as follows:
\begin{equation}
	\frac{\tau_{k+1}}{\tau_{k}} = \sqrt[3] {\frac{F\left(\boldsymbol{w}_{k+1}\right)}{F\left(\boldsymbol{w}_{k}\right)}}, \ \ \frac{s_{k+1}}{s_k} = \sqrt[3] {\frac{F\left(\boldsymbol{w}_{k}\right)}{F\left(\boldsymbol{w}_{k+1}\right)}}.
\end{equation}
Also, we can write them in terms of initial values $F\left(\boldsymbol{w}_{0}\right)$, $\tau_0$, and $s_0$ as follows:
\begin{equation}
	\tau_{k} = \sqrt[3] {\frac{F\left(\boldsymbol{w}_{k}\right)}{F\left(\boldsymbol{w}_{0}\right)}}  \tau_0, \ \ s_{k} = \sqrt[3] {\frac{F\left(\boldsymbol{w}_{0}\right)}{F\left(\boldsymbol{w}_{k}\right)}} s_0.
	\label{eq:conclusive}
\end{equation}
Equation (\ref{eq:conclusive}) yields conclusive results about the profiles of $\tau_k$ and $s_k$ over the course of learning. It implies that as the learning proceeds and the loss value $F\left(\boldsymbol{w}_{k}\right)$ gets smaller, the value of $\tau_{k}$ should also decrease, while the value of $s_k$ needs to increase.

We view both the communication-computation trade-off and the communication-precision trade-off as two distinct exploration-exploitation trade-offs, respectively---see Fig. \ref{fig:portray}. This view/perspective provides the rationale behind the joint and dynamic adjustment of $\tau_{k}$ and $s_k$; it also immensely assists to elucidate our proposed scheme FFL. Our two distinct exploration-exploitation trade-offs are explained as follows: at the beginning of learning, when the loss value is large, higher $\tau_{k}$ can accelerate learning (optimization) via aggressive exploration. However, once the loss becomes small, a large $\tau_k$ would fail to exploit. For $s_k$, meanwhile, we start with low values and over time increase it because when the loss is large, even imprecise gradients can contribute to decreasing the loss effectively---facilitate exploration. At lower losses, however, more precise gradients should be used---for exploitation. Fig. \ref{fig:portray} portrays the exploration-exploitation trade-offs: as time passes, for $k$th, $(k+1)$th, and $(k+2)$th rounds, the values of $\tau_{k}$, $\tau_{k+1}$, and $\tau_{k+2}$ decrease from $3$ and $2$ to $1$. Meantime, for $s_k$, over time, the values of $s_k$, $s_{k+1}$, and $s_{k+2}$ increase from $5$ and $7$ to $9$. It can be seen that over time the vector of \textit{compressed local aggregated gradient} gets closer to \textit{local aggregated gradient} because the compression is becoming more precise---more exploitation.

\begin{algorithm}[!t]
	\SetAlgoLined
	The server broadcasts $\boldsymbol{w}_{0}$\;
	Workers receive and initialize $\boldsymbol{w}_{0}$\;
	\For{$k=1, \cdots, K$}{
		The server calculates $\tau_k$ and $s_k$ using (\ref{eq:conclusive})\;
		The server broadcasts $\tau_k$ and $s_k$\;
		\For{$j=1, \cdots, M$ {\normalfont \textbf{in parallel}} }{
		Workers receive $\tau_k$ and $s_k$\;
		\For{$\ell=1,2, \ldots, \tau_k$}{
			Workers compute $\boldsymbol{g}(\boldsymbol{w}_{k}^{j, \ell}; \xi_{j})$\;
			Workers update $\boldsymbol{w}_{k}^{j, \ell}$ as in (\ref{eq:local_update_eq})\;
		}
		Workers compute $\boldsymbol{g}(\boldsymbol{w}_{k}^{j})$ as in (\ref{eq:localup})\;
		Workers compress $\boldsymbol{g}(\boldsymbol{w}_{k}^{j})$ as in (\ref{eq:estimator}) and (\ref{formu:sixth})\;
		Workers transmit $\widehat{\boldsymbol{g}}(\boldsymbol{w}_{k}^{j})$ to the server\;
		}
		The server receives $\widehat{\boldsymbol{g}}(\boldsymbol{w}_{k}^{j})$'s from workers\;
		The server averages $\widehat{\boldsymbol{g}}(\boldsymbol{w}_{k}^{j})$'s as in (\ref{eq:globalave})\;
		The server updates the global model as in (\ref{eq:globalup})\;
		The server broadcast $\boldsymbol{w}_{k+1}$ to workers\;
	}
	\caption{FFL at the server and workers.}
	\label{alg:masBased}
\end{algorithm}

Algorithm \ref{alg:masBased} presents FFL at the server and workers. Interestingly, FFL is similar to Model-Agnostic Meta-Learning (MAML) \cite{finn} in three aspects: (i) in FFL, we have \textit{workers} who posses their own datasets (with different distributions) while in MAML, there are \textit{tasks} with their corresponding datasets. (ii) In FFL, we have \textit{local updates} that improve performances (loss/accuracy) of individual workers whereas in MAML, there are \textit{inner-loop updates} that do so for individual tasks. (iii) In FFL, in each round, we apply \textit{global gradient} to weights, which improves the overall performance of all workers while in MAML, the \textit{outer-loop update} does so for all of tasks. Meanwhile, FFL is dissimilar to MAML in three facets: (i) FFL compresses \textit{local aggregated gradients} while MAML does not. (ii) FFL dynamically adjusts the local updates whereas MAML keeps the number of inner-loop updates fixed. (iii) In FFL, we only have \textit{one} update with global gradient while MAML can have multiple outer-loop updates. This coincidental similarity between FFL and MAML is promising as MAML has shown considerable success in \textit{accelerating} adaptation/generalization to different environments/tasks and the objective of FFL is achieving \textit{fast} FL by balancing trade-offs between communication and computation/precision.

In practical applications, our proposed scheme (Algorithm \ref{alg:masBased}) is implemented in the following steps: (i) each worker sends a participation signal to the server that serves as a participation request/permission. (ii) When the server, via referring to its database, approves (the reliability/trustworthiness of) the worker, it sends the latest weights as well as the local update coefficient and sparsity budget to the worker. The local update coefficient determines the number of gradient computations before communication, while the sparsity budget defines the rate of compression at which the worker is supposed to compress the gradients prior to communication. This way, the worker is synchronized with the FL system. (iii) The workers then, in each round, continuously compute gradients as many times as the local update coefficient allows and those workers transmit the compressed gradients at the compression rate that is specified by the sparsity budget. (iv) The server, meanwhile, receives the compressed gradients from the workers and computes the average gradient with which it updates the global weights. Later, the server jointly optimizes the local update coefficient and sparsity budget (via Theorem 6) for the upcoming round based on the latest loss value. The updated weights as well as the optimal local update coefficient and sparsity budget are then broadcast to workers.

\section{Experiments} \label{iamsecfive}
We used the TensorFlow 2.x deep learning library to conduct the experiments for performance evaluation of the proposed scheme. Specifically, we compare FFL with the following two state-of-the-art schemes: (i) ADACOMM \cite{adacom}, which exploits the idea of dynamically adjusting $\tau_{k}$, yet the workers do not compress the gradients for transmission and (ii) ATOMO \cite{atomo}, which makes use of gradient compression, but without dynamically adjusting $s_k$ over different rounds nor benefiting from dynamically adjusting $\tau_{k}$. We examine aforementioned schemes on two learning tasks: a Fully connected Neural Network (FNN) with the architecture of [784, 400, 400, 10] on MNIST, and a Convolutional Neural Network (CNN) that is VGG16 on CIFAR10. In all simulation scenarios, learning rate is set to $0.01$ and SGD optimizer is used---a simple optimizer lest simulation results get obscured \cite{noniid}. For compression, as in \cite{atomo}, we adopt Singular Value Decomposition (SVD). However, for implementation efficiency, we use thin-SVD \cite{brand2006fast} that calculates only $s$ number of singular values of $\boldsymbol{A} \in \mathbb{R}^{m \times n}$ with time-complexity of $O(mns)$ for $s \leq \sqrt{\min (m,n)}$ as opposed to full-SVD with $O(mn \times \min (m,n))$. Empirical observations suggest that most of the singular values of gradient matrices are often (very) small, i.e., (very) close to zero \cite{atomo, bigMatrix, dgc}; hence, calculating a small number of singular values for gradient matrices, as it is done in thin-SVD, often suffices. Note that neither our analysis nor the proposed scheme is not limited to SVD, which is used only as an example of compression in the experiments.

\begin{figure}[!t]
	\begin{subfigure}{0.33\columnwidth}
		\centering
		\includegraphics[width=\linewidth]{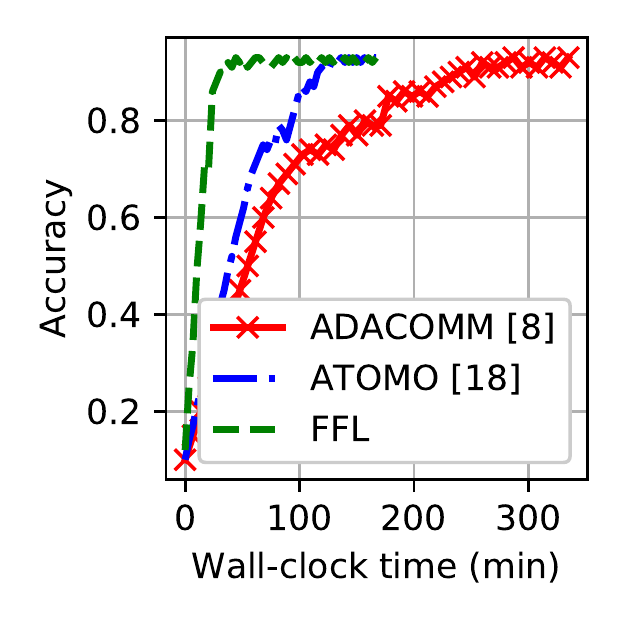}
		\caption{Accuracy}
		\label{fig:mnist32iidfullparti1}
	\end{subfigure}%
	\begin{subfigure}{0.33\columnwidth}
	\centering
	\includegraphics[width=\linewidth]{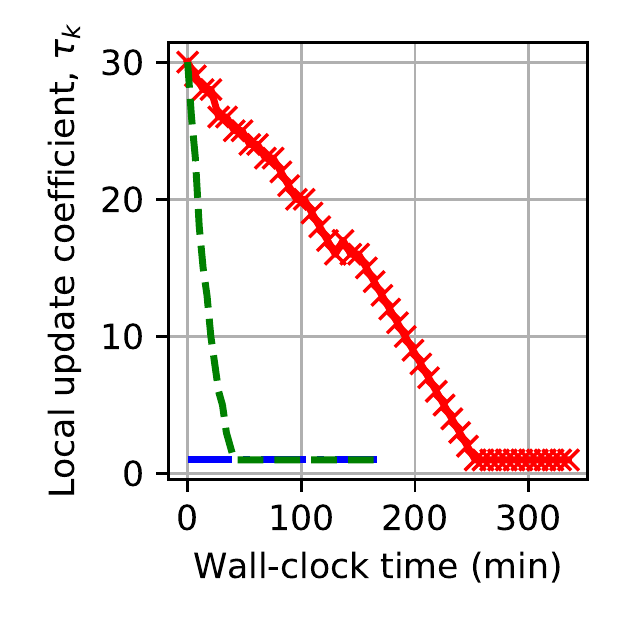}
	\caption{Local update, $\tau_k$}
	\label{fig:mnist32iidfullparti2}
	\end{subfigure}
	\begin{subfigure}{0.33\columnwidth}
		\centering
		\includegraphics[width=\linewidth]{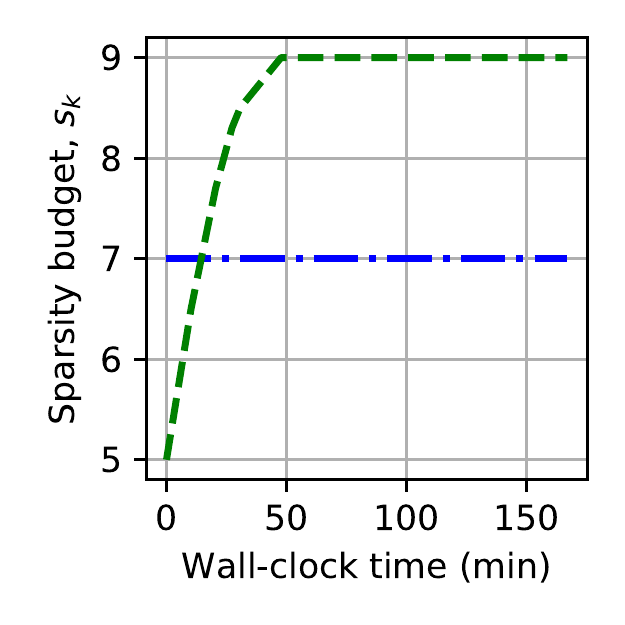}
		\caption{Sparsity budget, $s_k$}
		\label{fig:mnist32iidfullparti3}
	\end{subfigure}
	\caption{Accuracy, $\tau_k$, and $s_k$  values over wall-clock time for ADACOMM \cite{adacom}, ATOMO \cite{atomo}, and our proposed FFL with 32 workers for the FNN trained on MNIST. The dataset distribution is IID and all workers participate without momentum, while the server uses a momentum of 0.9. The uplink and downlink data rates are set equal to 100 $K$bps. Note that $\tau_k \in \{1, \cdots, 30 \}$ and $s_k \in [ 5,9]$.}
	\label{fig:mnist32iidfullparti}
\end{figure}

\begin{figure}[!t]
	\begin{subfigure}{0.33\columnwidth}
		\centering
		\includegraphics[width=\linewidth]{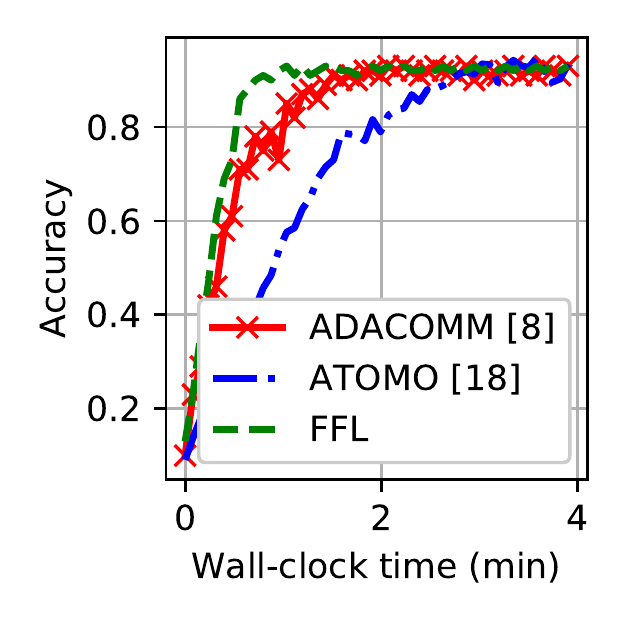}
		\caption{$10/10$ $M$bps}
		\label{fig:mnist32iidfullpartidr1}
	\end{subfigure}%
	\begin{subfigure}{0.33\columnwidth}
		\centering
		\includegraphics[width=\linewidth]{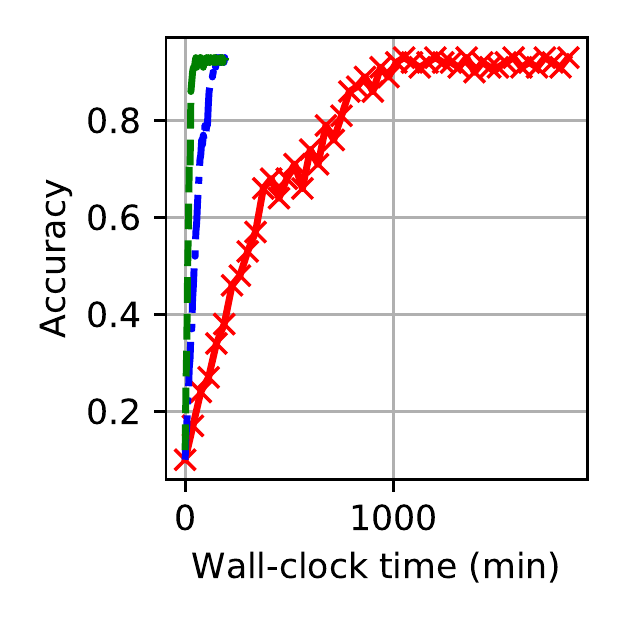}
		\caption{$10/100$ $K$bps}
		\label{fig:mnist32iidfullpartidr2}
	\end{subfigure}
	\begin{subfigure}{0.33\columnwidth}
		\centering
		\includegraphics[width=\linewidth]{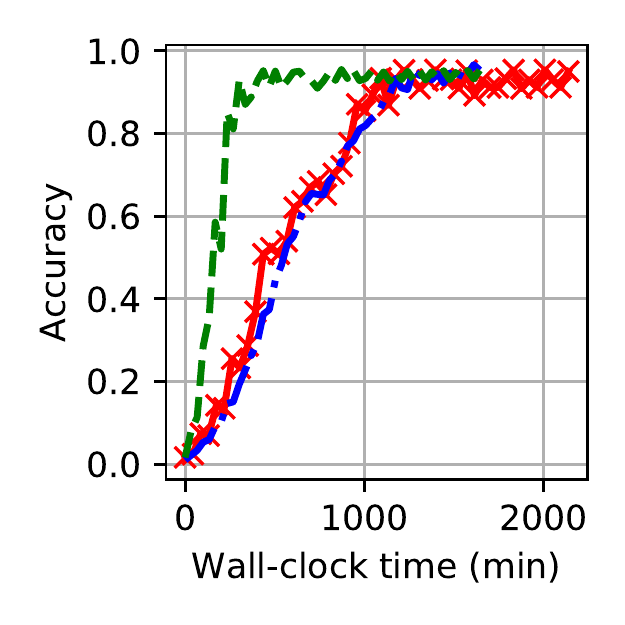}
		\caption{$100/10$ $K$bps}
		\label{fig:mnist32iidfullpartidr3}
	\end{subfigure}
	\caption{Accuracy for different uplink/downlink data rates. }
	\label{fig:mnist32iidfullpartidr}
\end{figure}

\begin{figure}[!t]
	\begin{subfigure}{0.33\columnwidth}
		\centering
		\includegraphics[width=\linewidth]{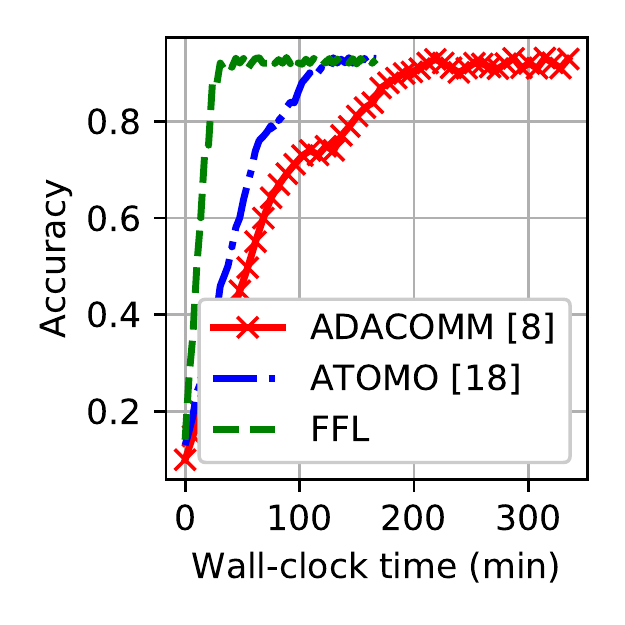}
		\caption{Momentum $0.2$}
		\label{fig:mnist32iidfullpartidrmom1}
	\end{subfigure}%
	\begin{subfigure}{0.33\columnwidth}
		\centering
		\includegraphics[width=\linewidth]{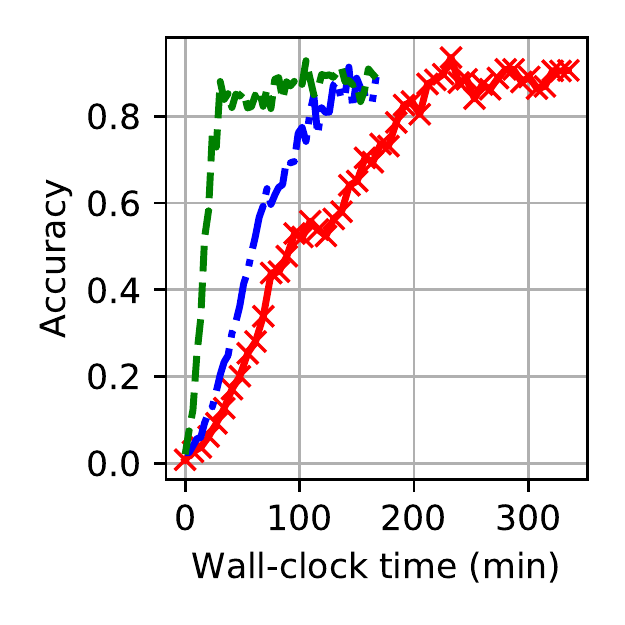}
		\caption{Momentum $0.6$}
		\label{fig:mnist32iidfullpartidrmom2}
	\end{subfigure}
	\begin{subfigure}{0.33\columnwidth}
		\centering
		\includegraphics[width=\linewidth]{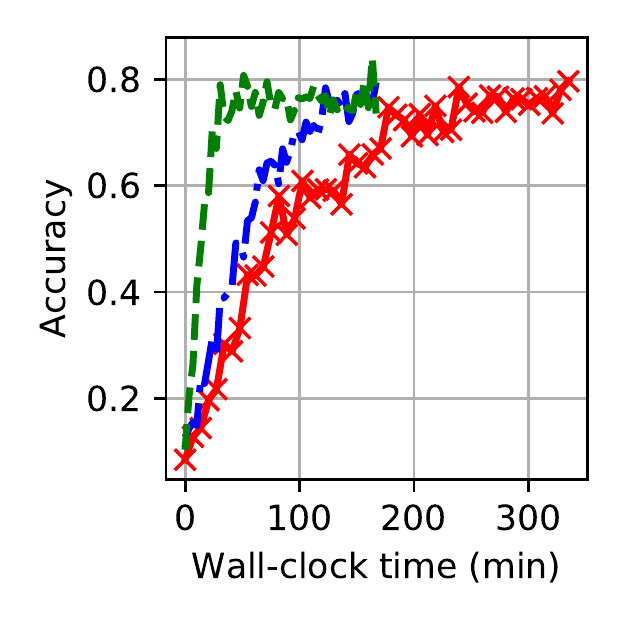}
		\caption{Momentum $0.8$}
		\label{fig:mnist32iidfullpartidrmom3}
	\end{subfigure}
	\caption{Accuracy for different momentums at workers. }
	\label{fig:mnist32iidfullpartidrmom}
\end{figure}

\begin{figure}[!t]
	\begin{subfigure}{0.33\columnwidth}
		\centering
		\includegraphics[width=\linewidth]{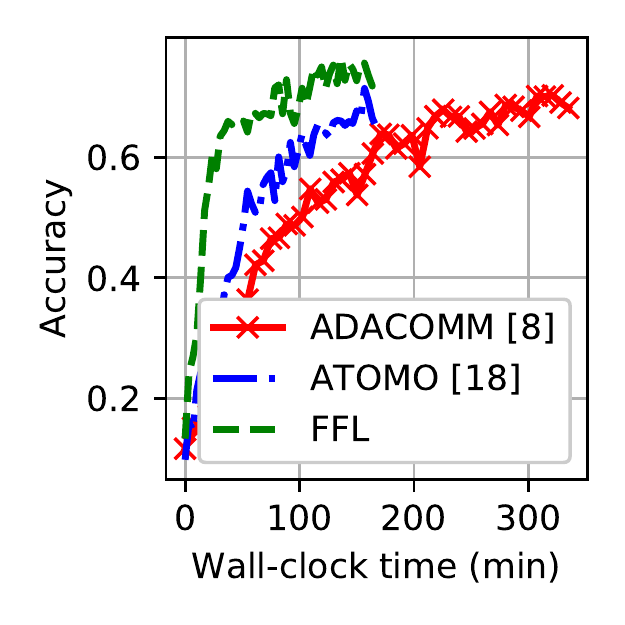}
		\caption{3 classes}
		\label{fig:mnist32iidfullpartidrnon1}
	\end{subfigure}%
	\begin{subfigure}{0.33\columnwidth}
		\centering
		\includegraphics[width=\linewidth]{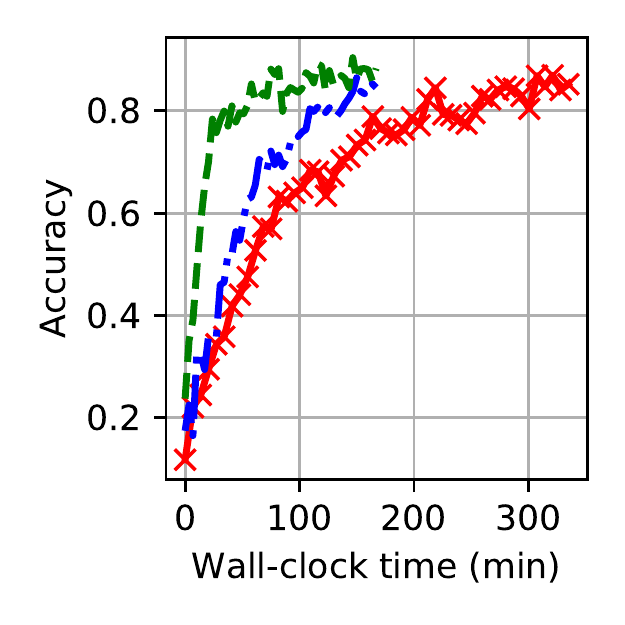}
		\caption{6 classes}
		\label{fig:mnist32iidfullpartidrnon2}
	\end{subfigure}
	\begin{subfigure}{0.33\columnwidth}
		\centering
		\includegraphics[width=\linewidth]{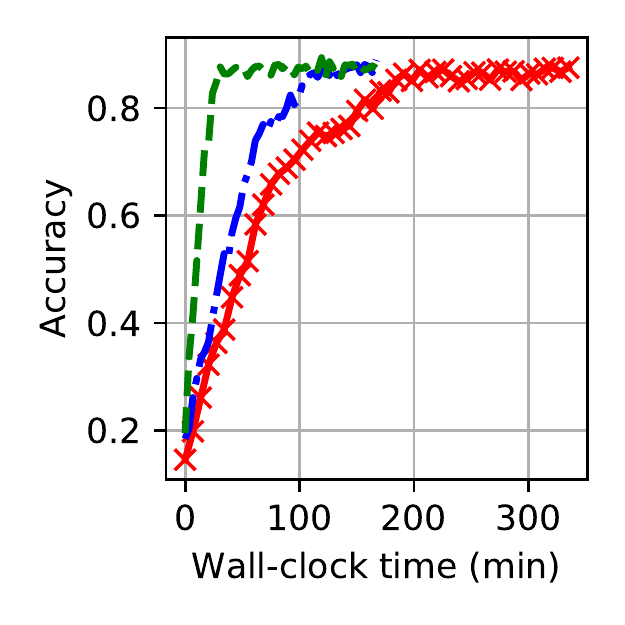}
		\caption{9 classes}
		\label{fig:mnist32iidfullpartidrnon3}
	\end{subfigure}
	\caption{Accuracy for different levels of non-IID-ness. }
	\label{fig:mnist32iidfullpartidrnon}
\end{figure}

\begin{figure}[!t]
	\begin{subfigure}{0.33\columnwidth}
		\centering
		\includegraphics[width=\linewidth]{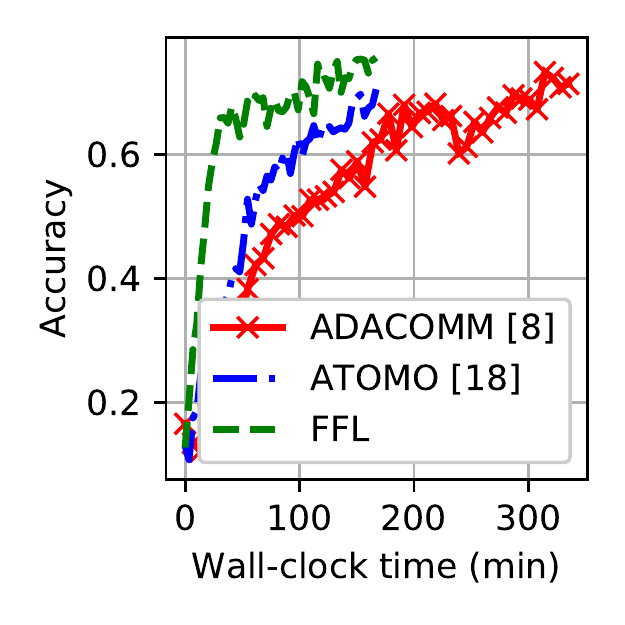}
		\caption{Probability of $0.7$}
		\label{fig:mnist32iidfullpartidrpar1}
	\end{subfigure}%
	\begin{subfigure}{0.33\columnwidth}
		\centering
		\includegraphics[width=\linewidth]{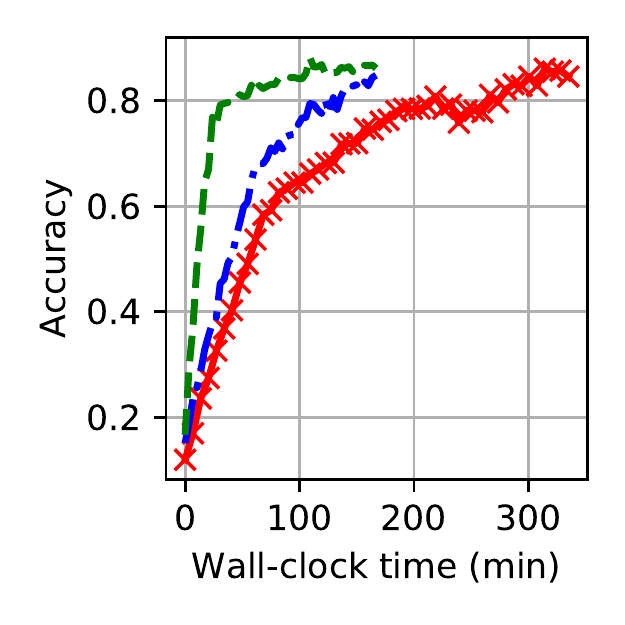}
		\caption{Probability of $0.4$}
		\label{fig:mnist32iidfullpartidrpar2}
	\end{subfigure}
	\begin{subfigure}{0.33\columnwidth}
		\centering
		\includegraphics[width=\linewidth]{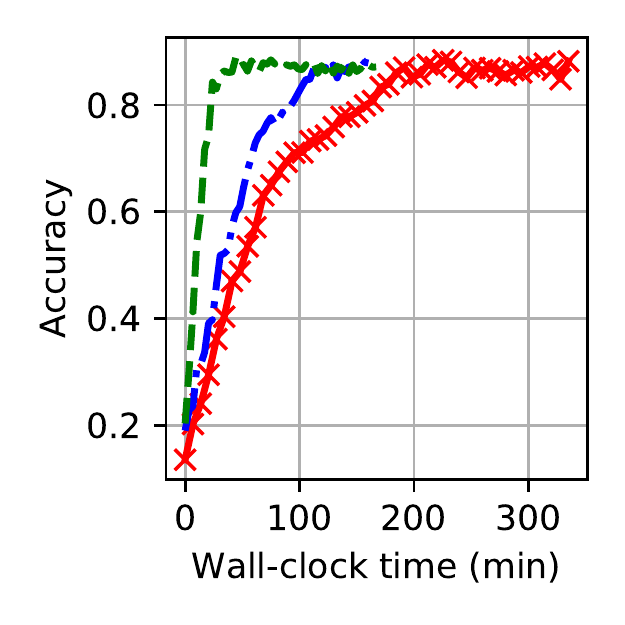}
		\caption{Probability of $0.1$}
		\label{fig:mnist32iidfullpartidrpar3}
	\end{subfigure}
	\caption{Accuracy for different probabilities of packet failure. }
	\label{fig:mnist32iidfullpartidrpar}
\end{figure}

\begin{figure}[!t]
	\begin{subfigure}{0.33\columnwidth}
		\centering
		\includegraphics[width=\linewidth]{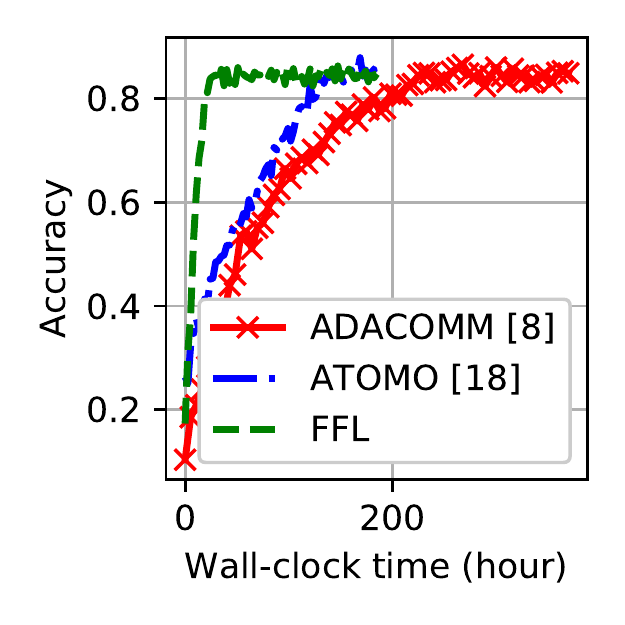}
		\caption{Accuracy}
		\label{fig:cifar32iidfullparti1}
	\end{subfigure}%
	\begin{subfigure}{0.33\columnwidth}
		\centering
		\includegraphics[width=\linewidth]{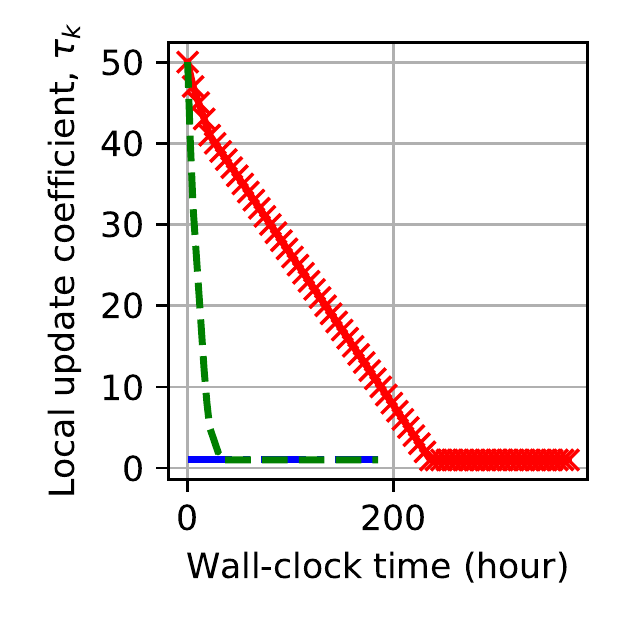}
		\caption{Local update, $\tau_k$}
		\label{fig:cifar32iidfullparti2}
	\end{subfigure}
	\begin{subfigure}{0.33\columnwidth}
		\centering
		\includegraphics[width=\linewidth]{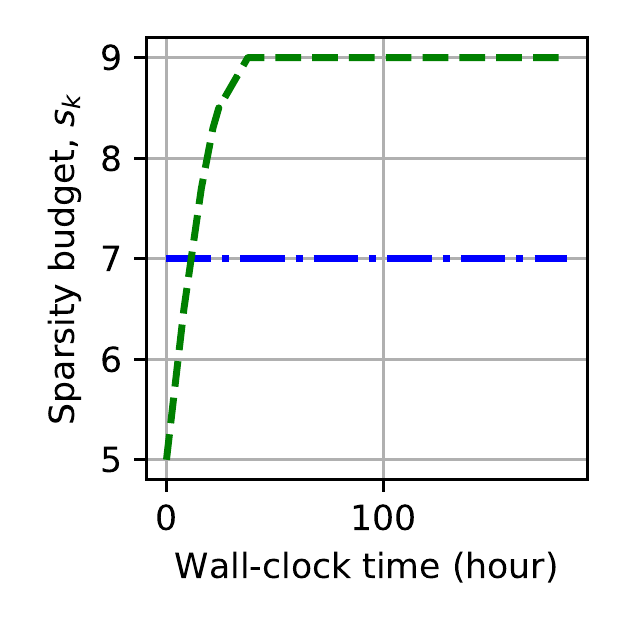}
		\caption{Sparsity budget, $s_k$}
		\label{fig:cifar32iidfullparti3}
	\end{subfigure}
	\caption{Accuracy, $\tau_{k}$, and $s_k$ for VGG16 on CIFAR10.}
	\label{fig:cifar32iidfullparti}
\end{figure}

The additional computation cost introduced by the SVD procedure reflects itself in the time axis of Figs. \ref{fig:mnist32iidfullparti}, \ref{fig:mnist32iidfullpartidr}, \ref{fig:mnist32iidfullpartidrmom}, \ref{fig:mnist32iidfullpartidrnon}, \ref{fig:mnist32iidfullpartidrpar}, and \ref{fig:cifar32iidfullparti}. Specifically, our scheme FFL uses the SVD compression while ADACOMM, which is one of our baselines, does not---ATOMO which is the other baseline does. This additional computation time of FFL that is spent for SVD compression causes computation part of each round of FFL to take slightly longer time compared to ADACOMM. Nonetheless, the SVD compression substantially reduces communication overheads, thereby saving considerable amount of time in compensation. Therefore, the overall time taken for a round is shorter in our scheme. This trade-off, which is spending time for computation in order to compress the gradients, is significantly time-saving because usually it is the communication time that is the bottleneck. Also, the amount of time spent for computation required for SVD compression is negligible compared to the gradient computation that includes performing $\tau$ number of local updates.  	 	 	 	

We plot variations of desired parameters (e.g., accuracy) against wall-clock time instead of FL rounds, epochs, or iterations. This is because unlike traditional centralized learning where different iterations take almost the same time and therefore the number of iterations reflects the speed of convergence, for FL that does not hold. The convergence speed of FL is determined by two factors: (i) the number of rounds and (ii) the time spent for each round that can \textit{significantly} differ over time because each round comprises (a) a downlink broadcast of global weights from the server to the workers with different link speeds (different delays), (b) local gradient computations that are again dependent on many factors such as the number of local updates and hardware heterogeneity, and finally,  (c) transmissions of the compressed gradients to the server by workers whose delays depend on compression ratios and uplink speeds. Hence, simply reporting the number of rounds is neither informative nor conclusive for the speed of convergence.

In experiments, for each worker, we first \textit{measure} the time it consumes for performing local computation and communication of gradients in each round. Then we select the maximum time consumed corresponding to the slowest worker as the time consumption of the round, because it is the slowest worker (the straggler) in each round that determines the time consumption of that round. We used the $time()$ function of $time$ package available in Python programming language for measuring the time spent for training process (local computation) and communication of various workers.

We explore the performance results of the three schemes along \textit{seven} distinct dimensions which characterize our learning environment: (i) Neural Network (NN) model, (ii) dataset, (iii) number of workers, (iv) uplink/downlink data rates, (v) momentum at workers, (vi) non-IID-ness, and (vii) noisy channels with packet failure. Throughout simulations, the data distribution among workers is balanced and non-overlapping. We adopt reasonable mini-batch sizes of 64 and 128 for MNIST and CIFAR10, respectively \cite{noniid}. In FL, it is crucial to adopt reasonable mini-batch sizes: because on one extreme, a too small mini-batch size cripples/prolongs the learning process due to high-variance SGD gradients---demanding even more communication overheads/rounds which in turn further slow the convergence \cite{noniid, atomo}. Recall that the variance of SGD gradients (denoted by $\sigma_2$ in this paper) is inversely proportional to the mini-batch size, and therefore the smaller the mini-batch size, the more high-variance/imprecise the gradients are \cite{adacom, cooperativesgd}\footnote{The main cause of high-variance gradients is having a non-representative mini-batch: a too small mini-batch size or a poorly shuffled mini-batch. Having a representative mini-batch to reduce the variance of gradients is necessary for acceleration of convergence \cite{atomo} both in general, and particularly for FL because in FL the learning process is distributed; this makes it more prone to gradient conflict/discrepancy and poor convergence. To secure representative-ness, a large (well-shuffled) mini-batch is vital.}. However, we should not overshoot: the size of the mini-batch must be just as large as to ensure that the mini-batch is fairly representative. As soon as representative-ness of the mini-batch is achieved, increasing the mini-batch size further only consumes dispensable resources---the other extreme. Because a larger mini-batch size proportionately demands larger memory and more computation which might be unaffordable particularly for resource-constranied IoT nodes and edge-devices \cite{noniid}. Interestingly, determining the mini-batch size itself involves balancing an inherent communication-computation trade-off: a too small mini-batch size (which translates to too little computation per communication) prolongs/cripples the learning, whereas an unnecessarily large mini-batch size (too much computation per communication) slows the learning.

In our experiments, the supported instantaneous communication data rate (more precisely, the instantaneous channel capacity) is time-invariant. Specifically, in the $t$-th time slot, the supported instantaneous communication data rate $R_t$ (bps) is given by
\begin{equation}
	R_t = W \log_2 \left(1 + \frac{ |h_t|^2 P_t}{\sigma^2} \right) 
\end{equation}
where $W$ is the assigned bandwidth (Hz), $P_t$ is the transmission power (watts) in the $t$-th time slot, $h_t$ is the channel coefficient in the $t$-th time slot, and $\sigma^2$ is the noise power (watts). In our experiments, we assume that the data rate $R_t$ is fixed throughout the course of learning by employing power control. For all data rates $R_t$ adopted in our experiments such as $10K$bps, $100K$bps, and $10M$bps, the bandwidth $W$ is constant, equal to $1M$Hz in all cases, channel coefficient $h_t$ varies among workers, and noise power is also kept constant at $1mW$. The transmission power $P_t$ is controlled so that the received signal to noise ratio, $|h_t|^2 P_t / \sigma^2$ yields the required data rate $R_t$.

Experiment results of the FNN on MNIST are shown in Fig. \ref{fig:mnist32iidfullparti} for 32 workers. Figs. \ref{fig:mnist32iidfullparti1}, \ref{fig:mnist32iidfullparti2}, and \ref{fig:mnist32iidfullparti3} show the values of accuracy, $\tau_k$, and $s_k$, respectively. Fig. \ref{fig:mnist32iidfullparti1} shows that, compared to the other two schemes, FFL achieves higher accuracies faster in terms of wall-clock time. In Fig. \ref{fig:mnist32iidfullparti2}, since ATOMO does not use local updates, it is set to one. The values of $s_k$ are plotted in Fig. \ref{fig:mnist32iidfullparti3} except for ADACOMM because it does not use gradient compression, whereas ATOMO does, but at a fixed $s_k$. Unlike these two schemes, FFL dynamically adjusts the value of $\tau_{k}$ and $s_k$ over time: for $\tau_{k}$, FFL starts with higher values to enable workers to perform an aggressive exploration. However, once the accuracy rises, the exploitation begins with lower $\tau_{k}$. For $s_k$, FFL starts with low values and over time increases because when the accuracy is small, even coarse gradients (i.e., highly compressed gradients due to low $s_k$) can still contribute to improving the accuracy of the model effectively. At higher accuracies, however, finer gradients are used for exploitation. Although all schemes in the figure perform 1560 rounds, ADACOMM takes twice as much time as others to converge because FFL and ATOMO are compressing their gradients in the uplink at a compression ratio of about $1/50$ associated with $s_k \in [ 5,9]$ that results in making the uplink communication time negligible compared to the downlink. Results of Fig. \ref{fig:mnist32iidfullparti} carry over to scenarios with different number of workers such as 4, 8, 16, and 64; we, therefore, avoid discussing those results again. In the rest of scenarios, hyperparameters are set the same as in Fig. \ref{fig:mnist32iidfullparti}; otherwise, it is mentioned.

In Fig. \ref{fig:mnist32iidfullpartidr}, we investigate three learning scenarios whose uplink and downlink data rates are different. First, in Fig. \ref{fig:mnist32iidfullpartidr1}, we consider high uplink/downlink communication data rates that equal to $10$ $M$bps---a relatively high data rate so as to dwarf the importance of compression. In such a case, not only compression is not that important in reducing the overall delay but also it can be detrimental due to its introduction of a compression time/delay. Indeed, here computation has become more of a bottleneck than communication. Expectedly, in Fig. \ref{fig:mnist32iidfullpartidr1}, ADACOMM is outperforming ATOMO despite its employment of compression. In this case, FFL approaches the performance of ADACOMM, even though FFL benefits from compression. Since employing compression is no longer the primary contributor in reducing the convergence time; what here plays the role instead is whether a scheme is using local updates or not. However, FFL is still outperforming ADACOMM due to a slight benefit it is taking from compression. Albeit, when uplink/downlink data rates approach infinity, ADACOMM would grow closer to FFL because the communication time of each round would approach zero and then it would not matter whether we compress gradients or not. In Fig. \ref{fig:mnist32iidfullpartidr2}, realistic values of uplink/downlink data rates ($10/100$ $K$bps) are chosen; here uplink communication is the bottleneck. In this case, the schemes which compress the gradients in the uplink win the race. Also, it can be seen that FFL has a slight advantage over ATOMO thanks to local updates. Conversely, in Fig. \ref{fig:mnist32iidfullpartidr3}, the downlink communication is the bottleneck which renders uplink compression futile. Consequently, ATOMO, despite its compression in the uplink, exhibits the same performance as ADACOMM; nonetheless, FFL outperforms both.

In Fig. \ref{fig:mnist32iidfullpartidrmom}, the impacts of using momentum (block momentum) at the workers are studied: we assess three values for momentum. In Fig. \ref{fig:mnist32iidfullpartidrmom1}, we notice that a small value for momentum at workers accelerates the learning compared to Fig. \ref{fig:mnist32iidfullparti1} where no momentum was used. Nevertheless, for higher values of momentum, the learning process starts to destabilize and yields lower accuracies, near $89\%$ and $73\%$ in Figs. \ref{fig:mnist32iidfullpartidrmom2} and \ref{fig:mnist32iidfullpartidrmom3}. This is because higher momentums at workers contribute in increasing the discrepancy among local gradients particularly because the dataset distribution is non-overlapping, which aggravates \textit{gradient conflict} among workers. The same behavior was observed in \cite{noniid} where high momentum crippled the learning. Yet, FFL is still more tolerant of higher momentums at workers than others.

Investigating non-IID data distribution over workers is demonstrated in Fig. \ref{fig:mnist32iidfullpartidrnon} where to each worker is assigned only 3, 6, and 9 classes in Figs. \ref{fig:mnist32iidfullpartidrnon1}, \ref{fig:mnist32iidfullpartidrnon2}, and \ref{fig:mnist32iidfullpartidrnon3}, respectively. Assigning 9 classes to workers (close to the IID scenario) does not inflict any noticeable harm, while for 6 classes, the baselines exhibit minor deterioration in performance; nonetheless, FFL is still on top. Finally, for 3 classes the accuracies undergo a considerable drop to $60\%$, which is unsurprising since non-IID learning is still an open problem even in centralized machine learning, let alone FL \cite{scafold}.

Performances for noisy channels with packet failure are examined in Fig. \ref{fig:mnist32iidfullpartidrpar} where at each global update only a percentage of the entire gradients reach the server; however, it is assumed that workers are still synchronized with the latest weights. The challenge of noisy channels with packet failure is that it can stall learning when only a set of \textit{non-representative gradients} reach the server and steer the learning away from the minimum and might even cause \textit{catastrophic forgetting} by directing the learning in the opposite direction. Results for noisy channels with packet failure probability of $0.1$, $0.4$, and $0.7$ are shown in Figs. \ref{fig:mnist32iidfullpartidrpar1}, \ref{fig:mnist32iidfullpartidrpar2}, and \ref{fig:mnist32iidfullpartidrpar3}.

Experiment results of the VGG16 on CIFAR10 are shown in Fig. \ref{fig:cifar32iidfullparti} for 16 workers. In this figure, compared to the previous results, FFL even more significantly outperforms the other two schemes. Numercally, whereas in the MNIST experiment in Fig. \ref{fig:mnist32iidfullparti}, FFL was $4\times$ faster than ATOMO, here in CIFAR10 experiment, FFL is $11 \times$ faster, because either of two techniques, local updates and gradient compression, becomes more necessary for larger NN models. Specifically, the larger the NN model, the more significant the role of dynamic local updates, which can accelerate the learning/optimization by dynamic adjustment of local update coefficients corresponding to the communication-computation trade-off. Meanwhile, when the NN model is small, the gradient constituents are also small matrices, which usually are not as low-rank as large matrices and therefore are not as suitable for compression, whereas large matrices tend to be highly low-rank \cite{bigMatrix, atomo, dgc}. Results of Fig. \ref{fig:cifar32iidfullparti} carry over to different number of workers such as 4 and 8. The other results observed for the FNN on MNIST such as effects of uplink/downlink data rates, momentum, non-IID-ness, and noisy channels with packet failure carry over to VGG16 on CIFAR10; hence, we avoid presenting them.

A summary of main results from the experiments are as follows: (i) as NN models scale, either of local update and gradient compression becomes more effective in accelerating learning, (ii) both dynamic adjustments of local update and gradient compression can be thought of as exploration-exploitation trade-offs: we start with high exploration (large $\tau_{k}$ and small $s_k$) and end with pure exploitation ($\tau_{k}=1$ and large $s_k$), (iii) FFL provides formulations to balance these trade-offs, and (iv) high momentums at \textit{workers}, non-IID-ness, and noisy channels with packet failure, which are the main causes of non-representative gradients inflict the least harm on FFL.

\section{Conclusion} \label{iamsecsix}
In this paper, our goal was to accelerate FL via minimizing learning error in a given wall-clock time with respect to local updates and gradient compression that correspond to trade-offs between communication and computation/precision, respectively. To this end, we first derived an upper bound of the learning error in a given wall-clock time considering the interdependency between the two variables: local update coefficient and sparsity budget. Based on this theoretical analysis, we then proposed an enhanced FL scheme, namely Fast FL (FFL), which jointly and dynamically adjusted the two variables to achieve \textit{fast} FL. In experiments, we demonstrated that FFL consistently achieved higher accuracies faster than similar schemes in the literature.

\ifCLASSOPTIONcaptionsoff
  \newpage
\fi

\bibliographystyle{IEEEtran}
\bibliography{references.bib}

\appendices
\renewcommand{\theequation}{\thesection.\arabic{equation}}
\setcounter{equation}{0}

\section{Proof of Lemma 2}
We determine the variance of the estimator in (\ref{eq:estimator}) starting from $\mathbb{E}_{ \{ e^{i}(\boldsymbol{w}_{k}^{j}) \} } \left[\| \widehat{\boldsymbol{g}}(\boldsymbol{w}_{k}^{j})-\boldsymbol{g}(\boldsymbol{w}_{k}^{j})\|^2 \right]$ as follows:
\begin{align}
&\mathbb{E}_{ \{ e^{i}(\boldsymbol{w}_{k}^{j}) \} } \left[\| \widehat{\boldsymbol{g}}(\boldsymbol{w}_{k}^{j})-\boldsymbol{g}(\boldsymbol{w}_{k}^{j})\|^2 \right]  \notag \\
& =  \mathbb{E}_{ \{ e^{i}(\boldsymbol{w}_{k}^{j}) \} } \left[ \left\|\sum_{i=1}^{B} \left( \frac{\lambda^{i}(\boldsymbol{w}_{k}^{j}) e^{i}(\boldsymbol{w}_{k}^{j})}{p^{i}(\boldsymbol{w}_{k}^{j})} \right) \boldsymbol{a}^{i}(\boldsymbol{w}_{k}^{j}) \right. \right. \notag \\
& \left. \left. \qquad \qquad \qquad \qquad \qquad \qquad  - \lambda^{i}(\boldsymbol{w}_{k}^{j}) \boldsymbol{a}^{i}(\boldsymbol{w}_{k}^{j})
 \right\|^2 \right] \\
& = \mathbb{E}_{ \{ e^{i}(\boldsymbol{w}_{k}^{j}) \} } \left[ \left( \sum_{i=1}^{B} \lambda^{i}(\boldsymbol{w}_{k}^{j}) \boldsymbol{a}^{i}(\boldsymbol{w}_{k}^{j}) \right. \right. \notag \\
& \left.  \left( \frac{ e^{i}(\boldsymbol{w}_{k}^{j}) - p^{i}(\boldsymbol{w}_{k}^{j})}{p^{i}(\boldsymbol{w}_{k}^{j})} \right) \right) ^T  \times \left( \sum_{i=1}^{B} \lambda^{i}(\boldsymbol{w}_{k}^{j}) \boldsymbol{a}^{i}(\boldsymbol{w}_{k}^{j}) \right. \notag \\
& \qquad \qquad \qquad \qquad \qquad  \left. \left. \left( \frac{ e^{i}(\boldsymbol{w}_{k}^{j}) - p^{i}(\boldsymbol{w}_{k}^{j})}{p^{i}(\boldsymbol{w}_{k}^{j})} \right) \right) \right] \\
& =  \sum_{i=1}^{B} \lambda^{i}(\boldsymbol{w}_{k}^{j})^2 \| \boldsymbol{a}^{i}(\boldsymbol{w}_{k}^{j}) \|^2 \notag \\
&  \qquad \qquad \times \mathbb{E}_{ \{ e^{i}(\boldsymbol{w}_{k}^{j}) \} } \left[ \left(\frac{ e^{i}(\boldsymbol{w}_{k}^{j}) - p^{i}(\boldsymbol{w}_{k}^{j})}{p^{i}(\boldsymbol{w}_{k}^{j})} \right)^2 \right] \notag \\
& \quad +  \sum_{r,t; r \neq t}^{B} \lambda^{r}(\boldsymbol{w}_{k}^{j}) \lambda^{t}(\boldsymbol{w}_{k}^{j}) \langle \boldsymbol{a}^{r}(\boldsymbol{w}_{k}^{j}),\boldsymbol{a}^{t}(\boldsymbol{w}_{k}^{j}) \rangle \notag \\
& \qquad \qquad \times \mathbb{E}_{ \{ e^{i}(\boldsymbol{w}_{k}^{j}) \} } \left[ \left(\frac{ e^{r}(\boldsymbol{w}_{k}^{j}) - p^{r}(\boldsymbol{w}_{k}^{j})}{p^{r}(\boldsymbol{w}_{k}^{j})} \right) \right. \notag \\
& \qquad \qquad \qquad \qquad \left. \times \left(\frac{ e^{t}(\boldsymbol{w}_{k}^{j}) - p^{t}(\boldsymbol{w}_{k}^{j})}{p^{t}(\boldsymbol{w}_{k}^{j})} \right) \right]. \label{eq:lemma2}
\end{align}
\noindent Now, we need to determine $\mathbb{E}_{ \{ e^{i}(\boldsymbol{w}_{k}^{j}) \} } \left[ \left(\frac{ e^{i}(\boldsymbol{w}_{k}^{j}) - p^{i}(\boldsymbol{w}_{k}^{j})}{p^{i}(\boldsymbol{w}_{k}^{j})} \right)^2 \right]$ and $\mathbb{E}_{ \{ e^{i}(\boldsymbol{w}_{k}^{j}) \} } \left[ \left( \frac{ e^{r}(\boldsymbol{w}_{k}^{j}) - p^{r}(\boldsymbol{w}_{k}^{j})}{p^{r}(\boldsymbol{w}_{k}^{j})} \right) \left(\frac{ e^{t}(\boldsymbol{w}_{k}^{j}) - p^{t}(\boldsymbol{w}_{k}^{j})}{p^{t}(\boldsymbol{w}_{k}^{j})} \right) \right]$. Starting with the first one, we have:
\begin{align}
 \mathbb{E}_{ \{ e^{i}(\boldsymbol{w}_{k}^{j}) \} } & \left[ \left(\frac{ e^{i}(\boldsymbol{w}_{k}^{j}) - p^{i}(\boldsymbol{w}_{k}^{j})}{p^{i}(\boldsymbol{w}_{k}^{j})} \right)^2 \right] \notag \\ 
 & =  (1-p^{i}(\boldsymbol{w}_{k}^{j}))\times \left(\frac{ 0 - p^{i}(\boldsymbol{w}_{k}^{j})}{p^{i}(\boldsymbol{w}_{k}^{j})} \right)^2 \notag \\
 & \qquad   + p^{i}(\boldsymbol{w}_{k}^{j})\times \left( \frac{ 1 - p^{i}(\boldsymbol{w}_{k}^{j})}{p^{i}(\boldsymbol{w}_{k}^{j})} \right)^2\\
& = \left(\frac{ 1}{p^{i}(\boldsymbol{w}_{k}^{j})} -1 \right). \label{eq-lemma2-1}
\end{align}

\noindent The second one can be written as follows:
\begin{align}
&\mathbb{E}_{ \{ e^{i}(\boldsymbol{w}_{k}^{j}) \} } \left[ \left(\frac{ e^{r}(\boldsymbol{w}_{k}^{j}) - p^{r}(\boldsymbol{w}_{k}^{j})}{p^{r}(\boldsymbol{w}_{k}^{j})} \right) \right.  \times \left. \left(\frac{ e^{t}(\boldsymbol{w}_{k}^{j}) - p^{t}(\boldsymbol{w}_{k}^{j})}{p^{t}(\boldsymbol{w}_{k}^{j})} \right) \right] \\
 & =  \mathbb{E}_{ \{ e^{i}(\boldsymbol{w}_{k}^{j}) \} } \left[ \left(\frac{ e^{r}(\boldsymbol{w}_{k}^{j}) - p^{r}(\boldsymbol{w}_{k}^{j})}{p^{r}(\boldsymbol{w}_{k}^{j})} \right) \right] \notag \\
& \qquad \times \mathbb{E}_{ \{ e^{i}(\boldsymbol{w}_{k}^{j}) \} } \left[ \left( \frac{ e^{t}(\boldsymbol{w}_{k}^{j}) - p^{t}(\boldsymbol{w}_{k}^{j})}{p^{t}(\boldsymbol{w}_{k}^{j})} \right) \right].
\end{align}

\noindent We show that $\mathbb{E}_{ \{ e^{i}(\boldsymbol{w}_{k}^{j}) \} } \left[ \left(\frac{ e^{r}(\boldsymbol{w}_{k}^{j}) - p^{r}(\boldsymbol{w}_{k}^{j})}{p^{r}(\boldsymbol{w}_{k}^{j})} \right) \right]$ and $\mathbb{E}_{ \{ e^{i}(\boldsymbol{w}_{k}^{j}) \} } \left[ \left(\frac{ e^{t}(\boldsymbol{w}_{k}^{j}) - p^{t}(\boldsymbol{w}_{k}^{j})}{p^{t}(\boldsymbol{w}_{k}^{j})} \right) \right]$ are zero.

\begin{align}
& \mathbb{E}_{ \{ e^{i}(\boldsymbol{w}_{k}^{j}) \} }  \left[ \left(\frac{ e^{r}(\boldsymbol{w}_{k}^{j}) - p^{r}(\boldsymbol{w}_{k}^{j})}{p^{r}(\boldsymbol{w}_{k}^{j})} \right) \right] \notag \\
 & =   (1 -p^{r}(\boldsymbol{w}_{k}^{j})) \left( \frac{ 0 - p^{r}(\boldsymbol{w}_{k}^{j})}{p^{r}(\boldsymbol{w}_{k}^{j})} \right) + p^{r}(\boldsymbol{w}_{k}^{j}) \left( \frac{ 1 - p^{r}(\boldsymbol{w}_{k}^{j})}{p^{r}(\boldsymbol{w}_{k}^{j})} \right) \\
& = 0. \label{eq:lemma2-2}
\end{align}
In the same way, the second term is zero. Resuming with (\ref{eq:lemma2}), we have: 
\begin{align}
& \mathbb{E}_{ \{ e^{i}(\boldsymbol{w}_{k}^{j}) \} }  \left[\| \widehat{\boldsymbol{g}}(\boldsymbol{w}_{k}^{j})-\boldsymbol{g}(\boldsymbol{w}_{k}^{j})\|^2 \right]  =  \sum_{i=1}^{B} \lambda^{i}(\boldsymbol{w}_{k}^{j})^2 \| \boldsymbol{a}^{i}(\boldsymbol{w}_{k}^{j}) \|^2 \notag \\
& \qquad \times \mathbb{E}_{ \{ e^{i}(\boldsymbol{w}_{k}^{j}) \} } \left[ \left(\frac{ e^{i}(\boldsymbol{w}_{k}^{j}) - p^{i}(\boldsymbol{w}_{k}^{j})}{p^{i}(\boldsymbol{w}_{k}^{j})} \right)^2 \right] \notag \\
& \qquad \quad +  \sum_{r,t; r \neq t}^{B} \lambda^{r}(\boldsymbol{w}_{k}^{j}) \lambda^{t}(\boldsymbol{w}_{k}^{j}) \langle \boldsymbol{a}^{r}(\boldsymbol{w}_{k}^{j}),\boldsymbol{a}^{t}(\boldsymbol{w}_{k}^{j}) \rangle \notag \\
& \qquad \qquad \times \mathbb{E}_{ \{ e^{i}(\boldsymbol{w}_{k}^{j}) \} } \left[ \left(\frac{ e^{r}(\boldsymbol{w}_{k}^{j}) - p^{r}(\boldsymbol{w}_{k}^{j})}{p^{r}(\boldsymbol{w}_{k}^{j})} \right) \right. \notag \\
& \qquad \qquad \qquad \qquad \times \left. \left(\frac{ e^{t}(\boldsymbol{w}_{k}^{j}) - p^{t}(\boldsymbol{w}_{k}^{j})}{p^{t}(\boldsymbol{w}_{k}^{j})} \right) \right]. \label{eq:lemma2-3}
\end{align}

\noindent The second term in (\ref{eq:lemma2-3}) was shown to be zero in (\ref{eq:lemma2-2}); thus, we have:
\begin{align}
& \mathbb{E}_{ \{ e^{i}(\boldsymbol{w}_{k}^{j}) \} }  \left[\| \widehat{\boldsymbol{g}}(\boldsymbol{w}_{k}^{j})-\boldsymbol{g}(\boldsymbol{w}_{k}^{j})\|^2 \right]  =  \sum_{i=1}^{B} \lambda^{i}(\boldsymbol{w}_{k}^{j})^2 \| \boldsymbol{a}^{i}(\boldsymbol{w}_{k}^{j}) \|^2 \notag \\
 & \qquad \times \mathbb{E}_{ \{ e^{i}(\boldsymbol{w}_{k}^{j}) \} } \left[ \left(\frac{ e^{i}(\boldsymbol{w}_{k}^{j}) - p^{i}(\boldsymbol{w}_{k}^{j})}{p^{i}(\boldsymbol{w}_{k}^{j})} \right)^2 \right].
\end{align}

\noindent From (\ref{eq-lemma2-1}) we know $\mathbb{E}_{ \{ e^{i}(\boldsymbol{w}_{k}^{j}) \} } \left[ \left(\frac{ e^{i}(\boldsymbol{w}_{k}^{j}) - p^{i}(\boldsymbol{w}_{k}^{j})}{p^{i}(\boldsymbol{w}_{k}^{j})} \right)^2 \right] = \left(\frac{ 1}{p^{i}(\boldsymbol{w}_{k}^{j})} -1 \right)$. Also, $\| \boldsymbol{a}^{i}(\boldsymbol{w}_{k}^{j}) \|^2 =1$. Therefore, we can write:

\begin{align}
 \mathbb{E}_{ \{ e^{i}(\boldsymbol{w}_{k}^{j}) \} } \left[\| \widehat{\boldsymbol{g}}(\boldsymbol{w}_{k}^{j})-\boldsymbol{g}(\boldsymbol{w}_{k}^{j})\|^2 \right] \notag \\ =  \sum_{i=1}^{B} \lambda^{i}(\boldsymbol{w}_{k}^{j})^2 \left(\frac{ 1}{p^{i}(\boldsymbol{w}_{k}^{j})} -1 \right).
\end{align}

\noindent This completes the proof. 
\hfill $\square$

\setcounter{equation}{0}
\section{Proof of Theorem 3}
We want to find an upper bound for $\mathbb{E}_{ \{ \xi_j \},  \{ e^{i}(\boldsymbol{w}_{k}^{j}) \} } \Big[ \|\widehat{\boldsymbol{g}}(\boldsymbol{w}_{k})-\nabla F(\boldsymbol{w}_{k})\|^{2} \Big]$; thus, we start by writing:
\begin{align}
 \mathbb{E}_{ \{ \xi_j \},  \{ e^{i}(\boldsymbol{w}_{k}^{j}) \} } &  \Big[ \|\widehat{\boldsymbol{g}}(\boldsymbol{w}_{k})-\nabla F(\boldsymbol{w}_{k})\|^{2} \Big] \notag \\
 & =  \mathbb{E}_{ \{ \xi_j \},  \{ e^{i}(\boldsymbol{w}_{k}^{j}) \} } \Big[ \|(\widehat{\boldsymbol{g}}(\boldsymbol{w}_{k})-\boldsymbol{g}(\boldsymbol{w}_{k})) \notag \\
 & \quad +(\boldsymbol{g}(\boldsymbol{w}_{k}) - \nabla F(\boldsymbol{w}_{k}))\|^{2} \Big] \\
& \leq \mathbb{E}_{ \{ e^{i}(\boldsymbol{w}_{k}^{j}) \} } \Big[ \|\widehat{\boldsymbol{g}}(\boldsymbol{w}_{k})-\boldsymbol{g}(\boldsymbol{w}_{k})\|^{2} \Big] \notag \\
& \quad + \mathbb{E}_{ \{ \xi_j \} } \Big[\|\boldsymbol{g}(\boldsymbol{w}_{k}) - \nabla F(\boldsymbol{w}_{k})\|^{2} \Big]. \label{eq:theorem3-1}
\end{align}
\noindent An upper bound for the second term in (\ref{eq:theorem3-1}), following \cite{adacom}, is obtained as $\beta\|\nabla F(\boldsymbol{w}_{k})\|^{2}+\sigma$. For the first term, the upper bound is determined as follows: 
\begin{align}
&\mathbb{E}_{ \{ e^{i}(\boldsymbol{w}_{k}^{j}) \} }  \Big[\|\widehat{\boldsymbol{g}}(\boldsymbol{w}_{k})-\boldsymbol{g}(\boldsymbol{w}_{k})\|^{2} \Big] \notag \\ 
& = \mathbb{E}_{ \{ e^{i}(\boldsymbol{w}_{k}^{j}) \} } \left[ \left\|\frac{1}{M} \sum_{j=1}^{M} \left (\widehat{\boldsymbol{g}}(\boldsymbol{w}_{k}^{j})-\boldsymbol{g}(\boldsymbol{w}_{k}^{j}) \right ) \right\|^{2} \right].
\end{align}

\noindent From Lemma 2, we know that $\mathbb{E}_{ \{ e^{i}(\boldsymbol{w}_{k}^{j}) \} } \left[\| \widehat{\boldsymbol{g}}(\boldsymbol{w}_{k}^{j})-\boldsymbol{g}(\boldsymbol{w}_{k}^{j})\|^2 \right] = \sum_{i=1}^{B} \lambda^{i}(\boldsymbol{w}_{k}^{j})^2 \left(\frac{ 1}{p^{i}(\boldsymbol{w}_{k}^{j})} -1 \right)$. Also, from Lemma 3, we know that $\sum_{i=1}^{B} \lambda^{i}(\boldsymbol{w}_{k}^{j})^2 \left(\frac{ 1}{p^{i}(\boldsymbol{w}_{k}^{j})} -1 \right) = \frac{\sigma_{1,j}^{k}}{s_k}  + \sigma_{2,j}^{k}$. Thus, we continue:
\begin{align}
\mathbb{E}_{ \{ e^{i}(\boldsymbol{w}_{k}^{j}) \} } & \Big[\|\widehat{\boldsymbol{g}}(\boldsymbol{w}_{k})-\boldsymbol{g}(\boldsymbol{w}_{k})\|^{2} \Big] \notag \\
& \leq \frac{1}{M}  \sum_{j=1}^{M} \sum_{i=1}^{B} \lambda^{i}(\boldsymbol{w}_{k}^{j})^2 \left(\frac{ 1}{p^{i}(\boldsymbol{w}_{k}^{j})} -1 \right) \\
& =  \frac{1}{M} \left(  \sum_{j=1}^{M}  \frac{\sigma_{1,j}^{k}}{s_k} + \sigma_{2,j}^{k} \right). \label{eq:theorem3-oo}
\end{align}
Therefore, from (\ref{eq:theorem3-1}) and (\ref{eq:theorem3-oo}), we have:
\begin{align}
\mathbb{E}_{ \{ \xi_j \},  \{ e^{i}(\boldsymbol{w}_{k}^{j}) \} } & \Big[ \|\widehat{\boldsymbol{g}}(\boldsymbol{w}_{k})-\nabla F(\boldsymbol{w}_{k})\|^{2} \Big] \notag \\
 \leq & \beta\|\nabla F(\boldsymbol{w}_{k})\|^{2} + \frac{\sigma_1}{s_k} +\sigma_2 
\end{align}
\noindent where $\sigma_1 = \max_{k} \left(\frac{1}{M} \sum_{j=1}^{M}\sigma_{1,j}^{k} \right)$ and $\sigma_2 = \max_{k} \left( \frac{1}{M} \sum_{j=1}^{M} \sigma_{2,j}^{k}\right) + \sigma$.
\hfill $\square$

\setcounter{equation}{0}
\section{Proof of Theorem 5}
For $\psi_k(\tau_k, s_k)$ to be convex, its Hessian matrix must be positive semidefinite. We derive the Hessian matrix of $\psi_k(\tau_k, s_k)$ as follows:
\begin{equation}
	\boldsymbol{H} (\psi_k(\tau_k, s_k)) =  \begin{bmatrix}  2A\frac{\alpha s_k}{\tau_{k}^3} & -A\frac{\alpha}{\tau_k^2} -C \frac{\sigma_{1}}{s_k^2} \\ -A\frac{\alpha}{\tau_k^2} -C \frac{\sigma_{1}}{s_k^2} & 2B\frac{\sigma_{1}}{s_k^3} + 2C(\tau_{k} -1) \frac{\sigma_{1}}{s_k^3} \end{bmatrix}
	\label{eq:hess}
\end{equation}
where $A= \frac{2\left[F\left(\boldsymbol{w}_{0}\right)-F_{\rm inf}\right]}{\eta T_k}$, $B=\frac{\eta L}{M}$, and $C=\eta^2 L^2$. If we ensure that both the diagonal elements and the determinant are positive, then we have proven that the matrix is positive semidefinite which would in turn guarantee the convexity of $\psi_k(\tau_k, s_k)$. Clearly, the elements on the diagonal are positive; meanwhile for the determinant we have:
\begin{align}
	& 4 \alpha \sigma_{1} \left( \frac{AB}{\tau_{k}^3 s_k^2} + \frac{AC (\tau_k -1)}{\tau_{k}^3 s_k^2}  - \left( \frac{A^2 \alpha }{\sigma_{1} \tau_{k}^4} + \frac{C^2 \sigma_{1}}{ \alpha s_k^4} + \frac{2AC}{\tau_{k}^2 s_k^2} \right) \right) \\
	& = 4 \alpha \sigma_{1} \left( \frac{4AB}{\tau_{k}^3 s_k^2} + \frac{2AC }{\tau_{k}^2 s_k^2} - \left( \frac{A^2 \alpha }{\sigma_{1} \tau_{k}^4} + \frac{C^2 \sigma_{1}}{\alpha s_k^4} + \frac{4AC}{\tau_{k}^3 s_k^2} \right) \right) \\
	& = \frac{2AC}{\tau_k^2 s_k^2} \left(1 - \frac{2}{\tau_{k}} \right) + \frac{4AB}{\tau_{k}^3 s_k^2} - \left ( \frac{A^2 \alpha}{\sigma_{1} \tau_{k}^4} + \frac{C^2 \sigma_{1}}{\alpha s_k^4} \right ) \\ 
	& \geq \frac{4B}{\tau_{k}^3 s_k^2} - \left ( \frac{A \alpha}{\sigma_{1} \tau_{k}^4} + \frac{C^2 \sigma_{1}}{ A \alpha s_k^4} \right ) \label{eq:onetwo} \\
	& = \frac{4B}{\tau_{k}^3 s_k^2} - \left ( \frac{A \alpha}{\sigma_{1} \tau_{k}^4} + \frac{\eta^5 L^4 T_k \sigma_{1}}{2 \alpha \left[F\left(\boldsymbol{w}_{0}\right)-F_{\rm inf}\right] s_k^4} \right ) \label{eq:ignore} \\
	& \approx \frac{4B}{\tau_{k}^3 s_k^2} -  \frac{A \alpha}{\sigma_{1} \tau_{k}^4} \label{eq:threefour} \\
	& = \frac{1}{M\eta T_k \sigma_{1} s_k^2 \tau_k^4} \left (  2 \eta L T_k \sigma_{1} \tau_{k} - \alpha M s_k^2 (F\left(\boldsymbol{w}_{0}\right)-F_{\rm inf}) \right) \label{eq:lastmove} \\
	& \geq 0.
\end{align}
\noindent According to assumption (i), $\tau_k \geq 2$, that justifies the move to (\ref{eq:onetwo}). Assumptions (ii) and (iii) help to ignore the third term in (\ref{eq:ignore}). The move to (\ref{eq:lastmove}) was due to assumption (iv).
\hfill $\square$

\setcounter{equation}{0}
\section{Lemma 3}

\noindent \textbf{Lemma 3}\textbf{:} \textit{The difference between compressed gradient $\widehat{\boldsymbol{g}}(\boldsymbol{w}_{k}^{j})$ and uncompressed gradient $\boldsymbol{g}(\boldsymbol{w}_{k}^{j})$ for the $j$th worker is given by}
\begin{equation}
	\mathbb{E}_{ \{ e^{i}(\boldsymbol{w}_{k}^{j}) \} } \Big [ \| \widehat{\boldsymbol{g}}(\boldsymbol{w}_{k}^{j})-\boldsymbol{g}(\boldsymbol{w}_{k}^{j})\|^2 \Big ] = \frac{\sigma_{1, j}^{k}}{s_k} + \sigma_{2, j}^{k}
	\label{eq:lemm3}
\end{equation}	
\noindent \textit{where $\sigma_{1, j}^{k} = \sum_{i=1}^{B}  \left( \lambda^{i}(\boldsymbol{w}_{k}^{j}) \| \lambda(\boldsymbol{w}_{k}^{j}) \|_1 \right)$ and $\sigma_{2, j}^{k} = -\sum_{i=1}^{B} \lambda^{i}(\boldsymbol{w}_{k}^{j})^2$.} \\
\noindent \textit{Proof:} To derive the difference between the compressed gradient
$\widehat{\boldsymbol{g}}(\boldsymbol{w}_{k}^{j})$ and uncompressed gradient $\boldsymbol{g}(\boldsymbol{w}_{k}^{j})$ for the $j$th worker, using Lemma 2 and Theorem 2, we have:
\begin{align}
& \mathbb{E}_{ \{ e^{i}(\boldsymbol{w}_{k}^{j}) \} } \left[ \| \widehat{\boldsymbol{g}}(\boldsymbol{w}_{k}^{j})-\boldsymbol{g}(\boldsymbol{w}_{k}^{j})\|^2 \right] \notag  \\
&= \sum_{i=1}^{B} \lambda^{i}(\boldsymbol{w}_{k}^{j})^2 \left(\frac{ 1}{p^{i}(\boldsymbol{w}_{k}^{j})} -1 \right) \label{eq:lemma3-1} \\
& = \sum_{i=1}^{B} \lambda^{i}(\boldsymbol{w}_{k}^{j})^2 \left(\frac{\| \lambda(\boldsymbol{w}_{k}^{j}) \|_1}{\lambda^{i}(\boldsymbol{w}_{k}^{j})s_k} -1 \right) \label{eq:lemma3-2} \\
& = \frac{1}{s_k} \sum_{i=1}^{B}  \left( \lambda^{i}(\boldsymbol{w}_{k}^{j}) \| \lambda(\boldsymbol{w}_{k}^{j}) \|_1 \right) - \sum_{i=1}^{B} \lambda^{i}(\boldsymbol{w}_{k}^{j})^2 \label{eq:lemma3-3} \\
& = \frac{\sigma_{1, j}^{k}}{s_k}  + \sigma_{2, j}^{k} \label{eq:lemma3-4}
\end{align}
where $\sigma_{1, j}^{k} = \sum_{i=1}^{B}  \left( \lambda^{i}(\boldsymbol{w}_{k}^{j}) \| \lambda(\boldsymbol{w}_{k}^{j}) \|_1 \right)$ and $\sigma_{2, j}^{k} = - \sum_{i=1}^{B} \lambda^{i}(\boldsymbol{w}_{k}^{j})^2$.
\hfill $\square$

\begin{IEEEbiography}[{\includegraphics[width=1.1in,height=1.3in,clip,keepaspectratio]{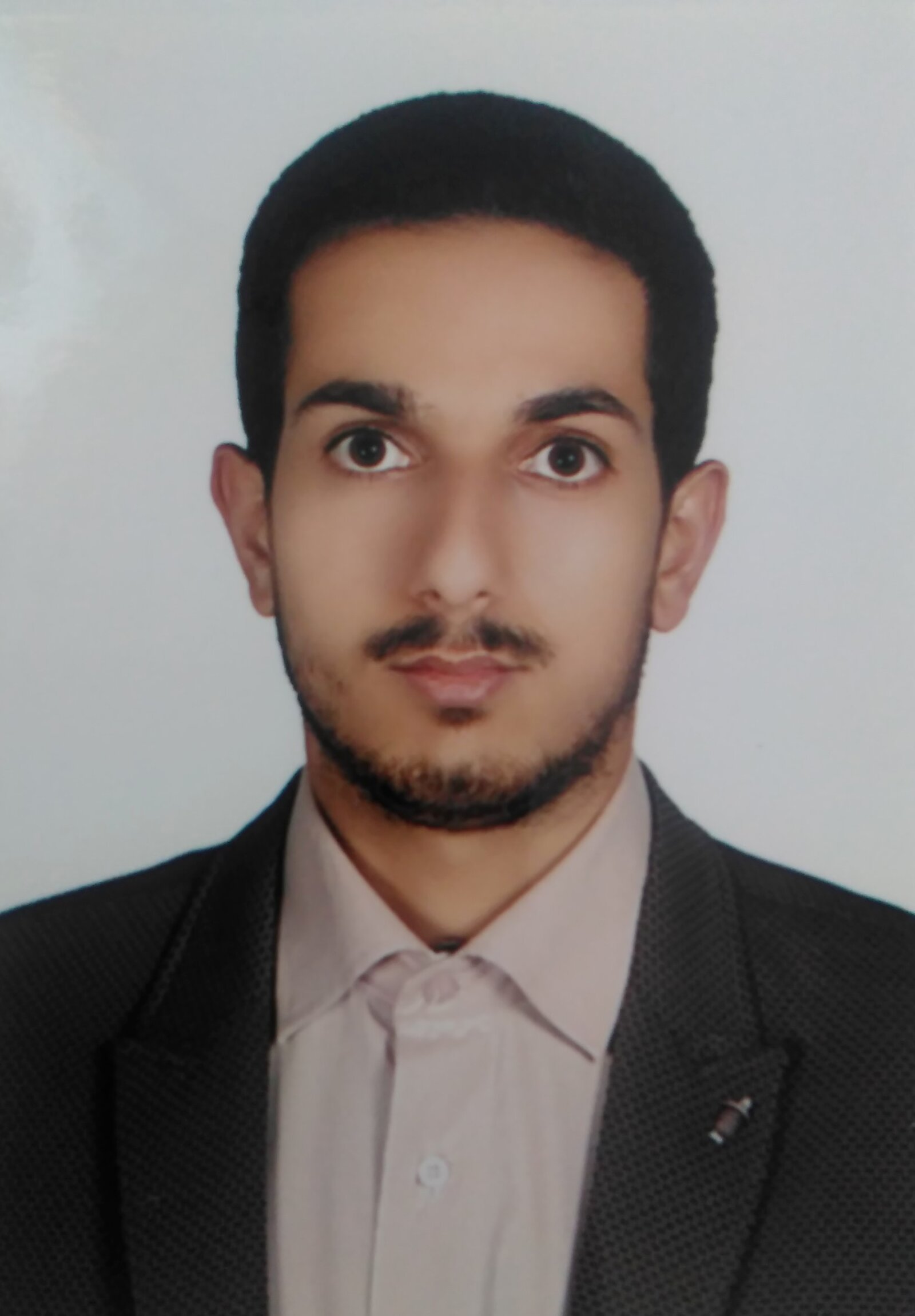}}]{Milad Khademi Nori} received his B.Sc. degree in electronics engineering as the 1st rank student from Semnan University, Semnan, Iran, in September 2016, and later his M.Sc. degree in digital electronics engineering from Amirkabir University of Technology, Tehran, Iran, in February 2019. Next, in August 2019, he joined the Electrical and Computer Engineering Department of Queen’s University, Kingston, ON, Canada, where he is currently a Ph.D. student working on the junction of artificial intelligence and (wireless) communication in the Wireless Artificial Intelligence Laboratory (WAI lab). In February 2016, he ranked 1st in the first phase of Iran’s national universities olympiad in the field of electrical engineering, and in November 2018, he was awarded in Iran’s national Information and Computer Technology (ICT) competition held by Huawei. His research interests include (Wireless) Communications, IoT, Machine Learning, and Federated Learning.
\end{IEEEbiography}

\begin{IEEEbiography}[{\includegraphics[width=1.1in,height=1.3in,clip,keepaspectratio]{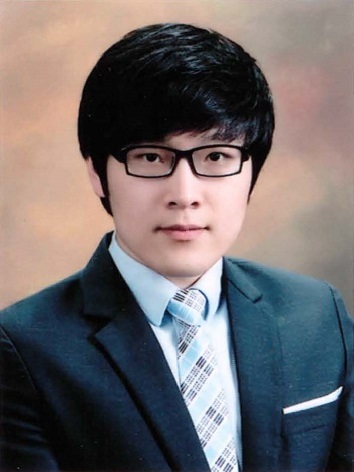}}]{Sangseok Yun}
received the Ph.D. degree in electrical engineering from the Korea Advanced Institute of Science and Technology (KAIST), Daejeon, South Korea, in 2018. In 2018, he was a Post-Doctoral Fellow at Korea Advanced Institute of Science and Technology (KAIST), Daejeon, South Korea, and visited Queen’s University, Kingston, Canada to conduct collaborative research. In 2019 and 2020, he was a Post-Doctoral Fellow at Queen’s University, Kingston, Canada. Since 2021, he has been with Pukyong National University, Busan, South Korea, where he is currently an Assistant Professor with the Department of Information and Communications Engineering. His research interests include theories and applications of deep learning, machine learning, wireless artificial intelligence, wireless communications, and physical layer security.
\end{IEEEbiography}

%\newpage

\begin{IEEEbiography}[{\includegraphics[width=1.05in,height=1.35in,clip,keepaspectratio]{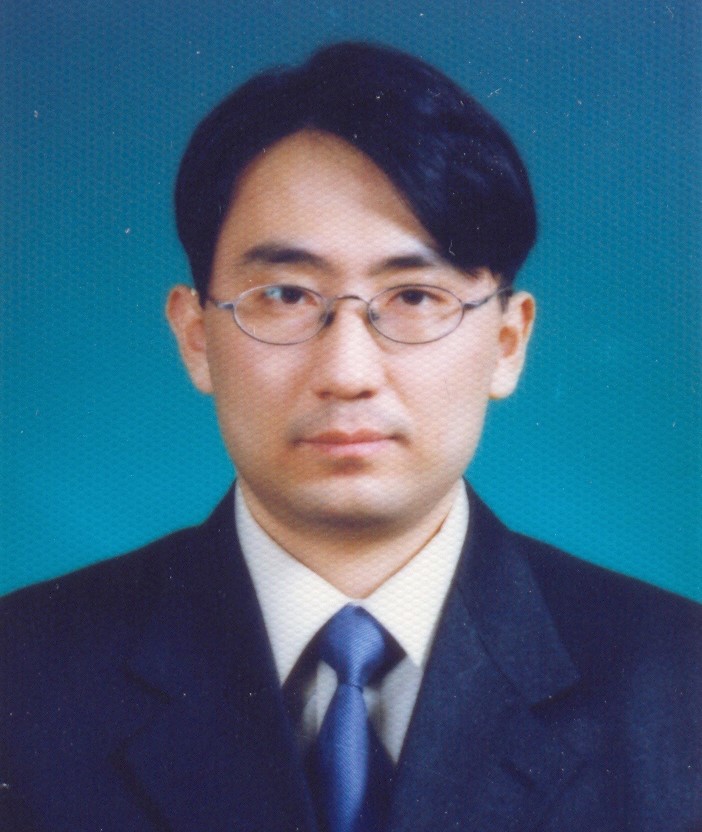}}]{Il-Min Kim} (SM’06) received the B.Sc. degree in electronics engineering from Yonsei University, Seoul, South Korea, in 1996, and the M.S. and Ph.D. degrees in electrical engineering from Korea Advanced Institute of Science and Technology (KAIST), Daejeon, South Korea, in 1998 and 2001, respectively. From October 2001 to August 2002, he was with the Department of Electrical Engineering and Computer Sciences, MIT, Cambridge, MA, USA, and from September 2002 to June 2003, he was with the Department of Electrical Engineering, Harvard University, Cambridge, MA, USA, as a Post-Doctoral Research Fellow. In July 2003, he joined the Department of Electrical and Computer Engineering, Queen’s University, Kingston, Canada, where he is currently a Professor. His current research interests include deep learning, reinforcement learning, continual learning, federated learning, Geo-AI, edge AI, deep learning for IoT and 6G. He has been an Editor for the IEEE TRANSACTIONS ON WIRELESS COMMUNICATIONS, the IEEE WIRELESS COMMUNICATIONS LETTERS, and the Journal of Communications and Networks (JCN).
\end{IEEEbiography}

\end{document}